\documentclass[11pt]{article}

\usepackage[final]{acl}

\usepackage{times}
\usepackage{latexsym}

\usepackage[T1]{fontenc}

\usepackage[utf8]{inputenc}

\usepackage{microtype}

\usepackage{inconsolata}

\usepackage{graphicx}

\usepackage{adjustbox}
\usepackage{amsmath}
\usepackage{booktabs}
\usepackage{algorithmic}
\usepackage{algorithm}

\title{Low-Resource Preference Adaptation of LLMs via Activation-Based Label Propagation}

\author{Alessio Galatolo \\
  Uppsala University \\
  \texttt{alessio.galatolo@it.uu.se} \\\And
  Meriem Beloucif \\
  Uppsala University \\
  \texttt{meriem.beloucif@lingfil.uu.se} \\}

\begin{document}
\maketitle
\begin{abstract}
Adapting large language models to user-specific preferences is often constrained by the cost of human annotation, making preference optimisation impractical in low-resource settings where preferences cannot be reliably labelled by LLMs themselves, e.g., due to cultural, subjective, or personalised contexts. In this paper, we investigate how language models encode preference information in their intermediate representations, finding that activations from chosen and rejected responses form distinct clusters across layers, even in pretrained models. Strikingly, this structure is strengthened by alignment on canonical datasets but erased when the target preferences differ from those the model was aligned on, suggesting aligned LLMs are poor judges for non-mainstream populations. Exploiting this structure, we propose training a lightweight linear probe on a few labelled preference pairs ($\leq$500) and using it to annotate large unlabelled datasets (50K+) for downstream preference optimisation. 
We systematically evaluate this approach across different datasets, preference optimisation methods and model scales and find that our method consistently outperforms direct training given the same annotation budget, and remains competitive against baselines trained on $50-100\times$ more labelled data in the majority of our settings.
\end{abstract}

\section{Introduction}

\begin{figure}[!t]
    \centering
    \includegraphics[width=\columnwidth]{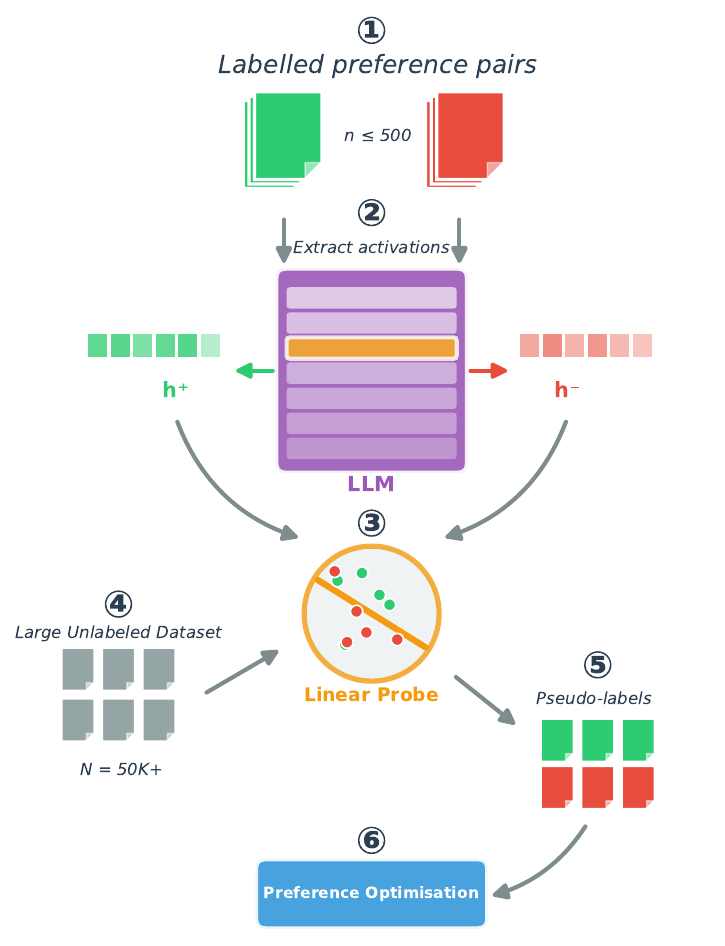}
    \caption{Illustration of our proposed pipeline. We start with (1) a small dataset with labelled preference pairs. Using the difference in activations of an LLM (2), we train a linear probe (3). We then use the probe to label a large corpus (4, 5) and do preference optimisation (6).}
    \label{fig:pipeline}
\end{figure}
Preference optimisation has emerged as a critical technique for aligning language models with human values and expectations \citep{ouyang2022training, rafailov2023direct}. Methods such as Direct Preference Optimisation (DPO) \citep{rafailov2023direct}, IPO ($\Psi$PO with identity mapping) \citep{azar2024general}, and their variants enable models to learn from human judgments about which responses are preferable. However, these methods typically require large datasets of labelled preference pairs, with annotation costs scaling linearly with dataset size. 

This requirement poses challenges for adapting models to specific user populations. Different communities, organisations, and cultural contexts exhibit distinct preferences \citep{kirk2024prism, sorensen2024value}: what constitutes a helpful or appropriate response varies substantially across groups. Obtaining large-scale annotations from each target population is often impractical, creating a barrier to personalised alignment.

A natural question arises: can we reduce the annotation burden while still achieving effective preference adaptation? One approach is to use an LLM (e.g., GPT-4 or newer) as an external judge \citep{zheng2023judging}, but these encode their own preference biases rather than those of the target population \cite{li-etal-2025-generation}. Reward model distillation requires substantial labelled data for training. Therefore, we propose a way to capture population-specific preferences with minimal supervision.

First, we investigate how LLMs encode preference-relevant information in their activations. Our results show that when processing chosen versus rejected responses, the resulting representations occupy different regions of activation space, forming clusters with distinct centroids even before preference optimisation (Figure~\ref{fig:clusters}). When using `traditional' preference optimisation datasets (i.e., datasets we expect to be similar to those used internally to train these models), this separation becomes more pronounced after supervised fine-tuning and preference optimisation, suggesting that instruction-following training implicitly organises representations according to response quality.

Building on this observation, we propose a simple pipeline (illustrated in Figure \ref{fig:pipeline}) for label-efficient preference adaptation:
\begin{enumerate}
    \item Collect a small set of labelled preference pairs from the target population ($n \approx$ 100-500).
    \item Train a linear probe (or similar small model) on model activations to distinguish chosen from rejected responses.
    \item Apply the probe to label a large corpus of unlabelled response pairs.
    \item Run standard preference optimisation using the probe-generated labels.
\end{enumerate}

Our experiments show that our approach consistently outperforms traditional training when given the same amount of annotations. When traditional training is given $50-100\times$ the amount of annotations, our method is still able to outperform two out of three baseline settings. Our findings lower the annotation barrier for preference adaptation, enabling effective adaptation in domains where human annotation budgets are severely constrained, including underrepresented and marginalised communities whose preferences have historically been underserved by large-scale models.

\section{Related Work}

\subsection{Preference Optimisation}
Reinforcement Learning from Human Feedback (RLHF) \citep{christiano2017deep, ouyang2022training} aligns language models with human preferences by training a reward model and optimising against it via reinforcement learning. Direct Preference Optimisation (DPO) \citep{rafailov2023direct} simplifies this by reparameterising the reward model as an implicit function of the policy, enabling direct optimisation on preference pairs. Subsequent work has proposed variants addressing limitations of DPO: IPO \citep{azar2024general} avoids overfitting via a different loss formulation, KTO \citep{ethayarajh2024kto} operates on unpaired examples using Kahneman-Tversky’s model of utility \cite{tversky1992advances}, and CPO \citep{xu2024contrastive} incorporates contrastive objectives. Our work studies how these methods respond to label noise, revealing substantial differences in robustness.

\subsection{Label-Efficient Alignment}
Prior work has explored reducing annotation requirements for alignment. Constitutional AI \citep{bai2022constitutional} uses model self-critique, while RLAIF \citep{lee2024rlaif} employs LLM judges for labelling. However, these approaches inherit the preferences of the labelling model rather than the target population. Further, unlike LLM-as-judge self-labelling, which we show collapses to chance for small models on subjective datasets (Section~\ref{sec:baselines}), our approach can operate directly on pretrained activations and does not require the model to articulate its own preference judgments. Reward model distillation \citep{fisch2025robust, askell2021general} transfers preferences but still requires substantial seed data. Recent work on weak-to-strong generalisation \citep{burns2024weak} studies supervision from less capable models. Our approach differs by exploiting structure in the model's own representations.

\subsection{Linear Probing in LLMs}
Linear probes have been used extensively to study representations in language models, revealing that models encode syntactic \citep{hewitt2019structural}, semantic \citep{tenney2019bert}, and factual \citep{meng2022locating} information in linearly accessible ways. Recent work has applied probing to detect sentiment \cite{hollinsworth-etal-2024-language} hallucinations \citep{azaria2023internal}, truthfulness \citep{marks2024geometry}, refusal behaviour \citep{arditi2024refusal}, model uncertainty \cite{wang2025response, dakhmouche2025can} and answer accuracy \cite{cencerrado2025answerneededpredictingllm}. A work that more closely aligns with ours is that of \citet{maiya-etal-2025-improving}, where linear probes are proposed as a replacement for LLMs-as-judges. However, their work specifically requires \textit{fine-tuned} LLMs and uses probes only for evaluation of performance.

We extend this line of work by showing that preference information is already encoded in pre-trained models and can thus be exploited for practical label propagation for fine-tuning. To the best of our knowledge, we are the first to use probes for label propagation at scale and to apply it for preference optimisation.

\section{Preference Geometry in Activations}
\label{sec:geometry}

We begin by characterising how preference information is organised in language model representations.

\subsection{Experimental Setup}

We analyse activations from different models of Llama 3.2 \cite{llama3}, Gemma 3 \cite{gemma3}, and Qwen 3 \cite{qwen3} at various training stages: pretrained, or after SFT and Preference Optimisation (PO). For each stage, we process preference pairs from HH-RLHF \cite{bai2022training}, UltraFeedback \cite{cui2023ultrafeedback}, and Nectar \cite{nectar}. On top of these `standard' datasets, we also analyse PRISM \cite{kirk2024prism} as an example of culturally-diverse preferences, oasst2\footnote{\url{https://huggingface.co/datasets/OpenAssistant/oasst2}} \cite{oasst2} for multilingual preferences and AfriSenti \cite{muhammad-etal-2023-afrisenti} for sentiment analysis in low-resource languages. For each sample, we extract the activations of positive and negative samples. The activations for a single sequence are then aggregated by either taking: (i) the mean of the completion tokens, (ii) only the token with maximum activations, or (iii) only the activations for the final token in the sequence.

\subsection{Cluster Structure}

\begin{figure*}[t]
    \centering
    \includegraphics[width=\textwidth]{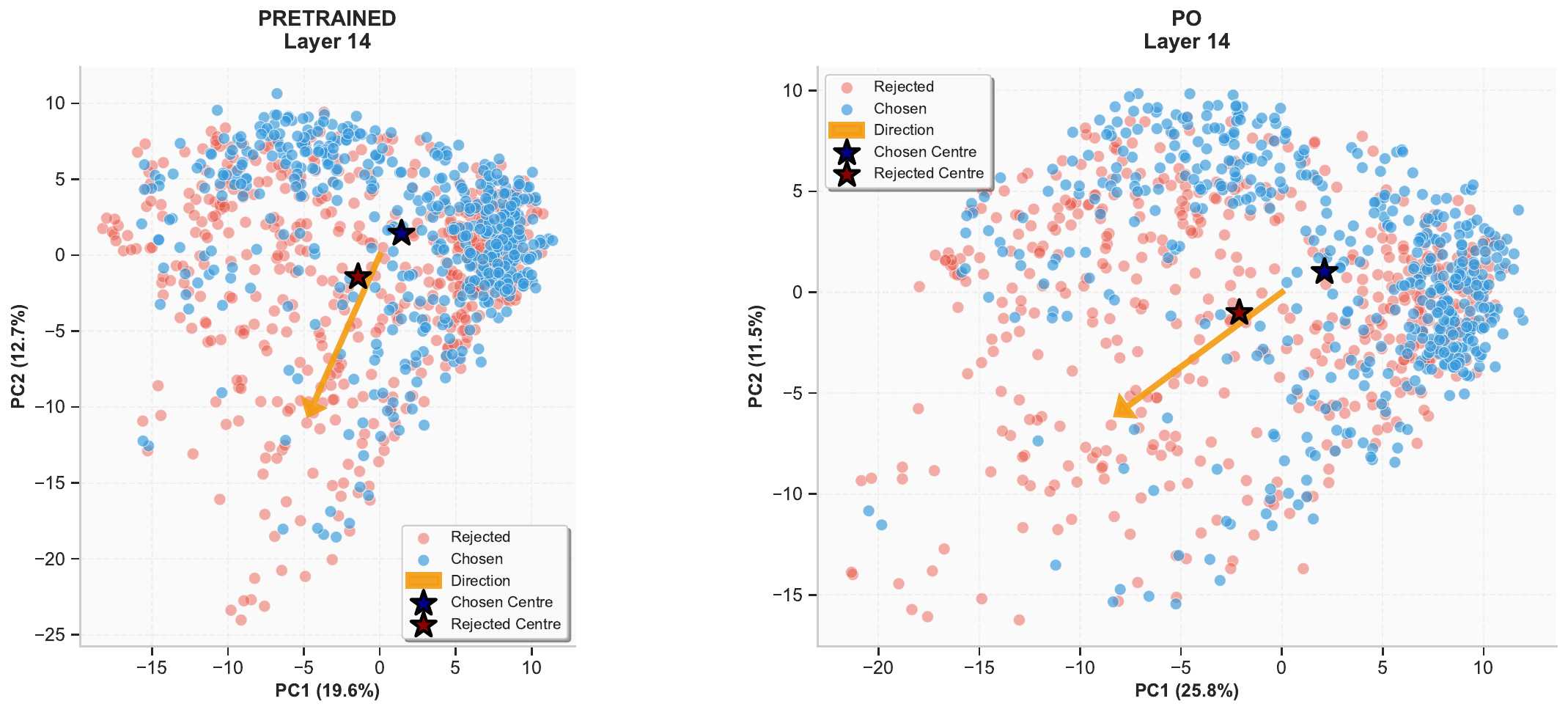}
    \caption{Activations from chosen (blue) and rejected (orange) responses form separable clusters using Qwen 3 0.6B and Nectar. The distance between centroids increases in PO compared to pretrained.}
    \label{fig:clusters}
\end{figure*}

We visualise activations from chosen and rejected responses using t-SNE \citep{van2008visualizing} and PCA \cite{pca}, noticing that, while chosen and rejected clusters mostly overlap, their centroids are quite distinct. This separation is present even in pretrained models but becomes more pronounced (up to 40\% more in distance) after SFT and PO. We show in Figure~\ref{fig:clusters} one of the most prominent examples, obtained with Qwen 3 0.6B and the Nectar dataset. The pattern holds across model families and datasets, though the dataset has the largest effect on cluster separation: HH-RLHF, for instance, shows a less pronounced difference, consistent with its noisier annotations.

As a sanity check, we re-plot (and report in Appendix~\ref{app:sanity_check}) activations with chosen/rejected labels randomly permuted; the clusters collapse onto a shared centroid, confirming the separation reflects genuine preference structure rather than visualisation artefact.

\paragraph{Visual separation is statistically significant.}
To confirm that the observed separation is not an artefact of visualisation, we conduct a systematic battery of statistical tests across 9 settings (3 datasets $\times$ 3 model families). We project activations onto the first principal component (PC1) and apply a two-sample $t$-test, falling back to a Mann-Whitney $U$ test when normality assumptions are violated. In five out of nine settings we find $p \ll 0.001$; in three of the remaining four we find $p < 0.05$; only one combination fails to reach significance at the 0.05 level. 

To complement the univariate analysis, we compute Cohen's $d$ on the full-dimensional activation space, obtaining values ranging from 2.63 to 3.75 across all settings. By standard conventions, $d > 0.8$ is a `large' effect; our values exceed this threshold by a factor of $3$--$4\times$, indicating a very substantial separation between the chosen and rejected centroids. We additionally run a kernel two-sample test (MMD, RBF kernel with median-heuristic bandwidth, permutation-based $p$-values, $n{=}250$ per class) directly on the full-dimensional activations. This multivariate test corroborates the univariate finding: eight of nine settings reach significance at $p<0.05$, with the sole exception being Llama 3.2 tested with UltraFeedback ($p=0.11$). The MMD magnitude tracks the same ordering already observed for cluster separation and probe accuracy, with Nectar showing the largest effect and UltraFeedback/HH-RLHF the smallest (full results in Appendix~\ref{app:mmd}). These results confirm that the cluster structure visible in Figure~\ref{fig:clusters} reflects a statistically robust phenomenon.

\subsection{Accuracy and Layer-wise Analysis}

To quantify separability, we train linear probes at each layer and measure classification accuracy on held-out data. 

Firstly, we find that probe accuracy peaks in the middle-to-late layers, consistent with findings on other representation probing tasks \cite{gurnee2024language}. We also find that the accuracy of pretrained and PO models is mostly similar, with some families of models showing slightly higher accuracy for pretrained (e.g., Llama 3.2), while others showing slightly higher accuracy for PO (Gemma 3, Qwen 3). Similar to before, we find that the dataset has the biggest influence on probe accuracy, with Nectar peaking above 75\% while HH-RLHF plateauing around 60\%.

Note that, while probe accuracy of 60\% may appear low, preference annotation can be quite noisy. In HH-RLHF, the reported inter-annotator agreement is approximately 63\%, indicating that our probe approaches the ceiling of human consistency. This suggests the probe may capture a genuine preference signal rather than merely failing to learn the task. Full per-setting results with confidence intervals are reported in Appendix \ref{app:probe_acc}, alongside illustrations of probe accuracy across layers (\ref{app:layerwise}).

\paragraph{The effect is not an artefact of pretraining contamination.}
An important concern is whether the observed separation is an artefact of the model having memorised preference labels during pretraining. To control for this, we run an additional experiment using Llama 2 \cite{llama2} (released July 2023) paired with the Nectar dataset (released November 2023), ensuring a strict temporal separation between the model and the data. The results closely mirror those reported for other settings: peak probe validation accuracy is $78.3\% \pm 1.9$, the $t$-test on PC1 yields $p \ll 0.001$, and Cohen's $d = 3.73$. This demonstrates that the preference geometry in activations is a genuine property of how language models represent response quality, not an artefact of memorisation.

\paragraph{Results are stable across settings.}
Varying the aggregation method, i.e., mean of activations vs. last token vs. max, does not affect clustering nor probe performance. However, an important aspect of computing the mean is that it should be calculated solely over the completion, excluding the prompt. Averaging over the entire input leads to clusters that are no longer distinguishable.

\subsection{Alignment Erases Non-Canonical Preference Geometry}
\label{sec:culturally_divers}
\begin{figure*}[!t]
    \centering
    \includegraphics[width=\textwidth]{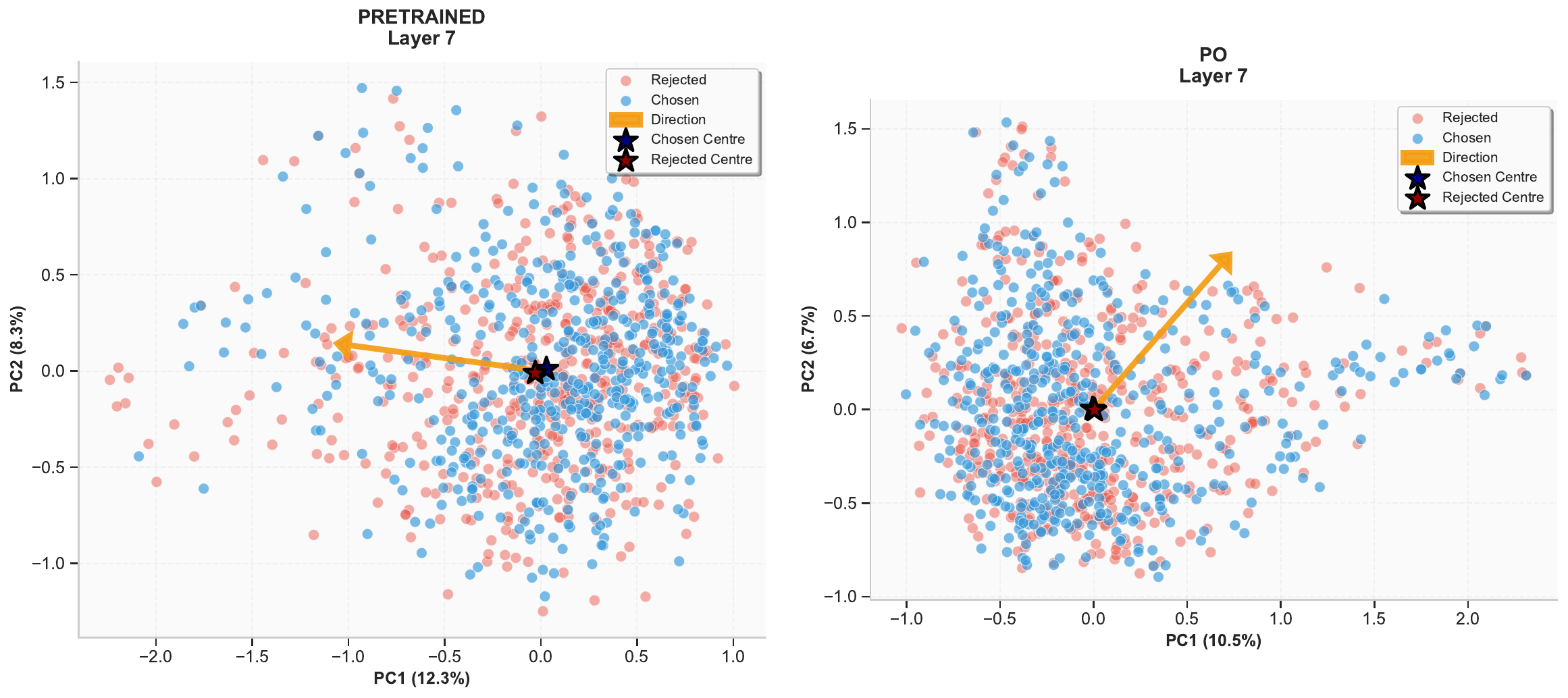}
    \caption{Activations from chosen (blue) and rejected (orange) responses form separable clusters using Llama 3.2 1B and PRISM. Distance between centroids \textit{decreases} in PO compared to pretrained.}
    \label{fig:prism_clusters}
\end{figure*}

The main application of our findings is in low-resource scenarios and culturally specific settings, where data is too scarce. We begin by analysing multilingual capabilities in preference optimisation with oasst2 and find that all models achieve 65-69\% accuracy across seven different languages (de, fr, es, zh, ru, it, jp).
We also expand our experiments to sentiment analysis in low-resource languages using the AfriSenti dataset, where we find high accuracy across most languages ($70+\%$, up to $84\%$ for Mozambique Portuguese), with Amharic scoring the lowest at $58-60\%$; we attribute this to the different writing system of the language. 

Finally, we analyse the activations of PRISM, a preference dataset collected from a diverse range of populations where each sample is paired with the annotator's cultural background. We begin by grouping conversations in PRISM by user location and age group, and we discover similar clustering and probe performance to those of `standard' preference datasets. A key difference here is that, while the pretrained model shows distinct centroids for chosen and rejected samples, this distinction is much reduced or not present \textit{at all} in the default SFT/PO version of the model (Figure \ref{fig:prism_clusters}). This strengthens the hypothesis that models aligned to a particular set of preferences are unable to distinguish between a \textit{different} set of preferences and are thus unfit to be used as judges or annotators \cite{li-etal-2025-generation}.

\section{Method}
\label{sec:method}

Exploiting the preference geometry presented in the previous Section, we propose a pipeline for label-efficient preference adaptation assuming access to:
\begin{itemize}
    \item A language model $\pi_\theta$ fine-tuned for instruction following (SFT model)
    \item A small labelled dataset $\mathcal{D}_L = \{(x_i, y_i^+, y_i^-)\}_{i=1}^{n}$ with $n \approx$ 10-500
    \item A large unlabelled dataset $\mathcal{D}_U = \{(x_j, y_j^a, y_j^b)\}_{j=1}^{N}$ with $N \gg n$
\end{itemize}

The goal is to adapt $\pi_\theta$ to the preferences represented in $\mathcal{D}_L$ by leveraging $\mathcal{D}_U$.

\subsection{Probe Training}

For each labelled example $(x, y^+, y^-)$, we extract from $\pi_\theta$ activations at layer $\ell$ and select those relative to the last token in the sequence:
\begin{align}
    h^+ &= \text{Activation}_\ell(\pi_\theta, [x; y^+]) \\
    h^- &= \text{Activation}_\ell(\pi_\theta, [x; y^-])
\end{align}

We train a linear probe $f_\phi(h) = \sigma(W h + b)$ to predict preference:
\begin{equation}
    \mathcal{L}_\text{probe} = -\sum_{i} \left[ \log f_\phi(h_i^+) + \log(1 - f_\phi(h_i^-)) \right]
\end{equation}

Here, the choice of $\ell$ can be either made through a validation set or with an `informed' guess. Based on the analysis in Section~\ref{sec:geometry}, we believe the best layers are those starting from the middle of the model, e.g., for a model with $L$ layers, layers $\lfloor L/2 \rfloor$ to $\lfloor 2L/3 \rfloor$.

\subsection{Label Propagation}

For each unlabelled pair $(x, y^a, y^b) \in \mathcal{D}_U$, we compute:
\begin{align}
    p^a &= f_\phi(\text{Activation}_\ell(\pi_\theta, [x; y^a])) \\
    p^b &= f_\phi(\text{Activation}_\ell(\pi_\theta, [x; y^b]))
\end{align}

We assign labels based on probe predictions:
\begin{equation}
    (y^+, y^-) = \begin{cases}
        (y^a, y^b) & \text{if } p^a > p^b \\
        (y^b, y^a) & \text{otherwise}
    \end{cases}
\end{equation}

\subsection{Preference Optimisation}

We apply standard preference optimisation methods (DPO, IPO, KTO, CPO) on the probe-labelled dataset. The key question is how each method responds to the noise introduced by imperfect probe predictions.

\section{Experiments}
\label{sec:experiments}

We conduct several experiments to validate the effectiveness of models trained on probe labels. The code for all the analysis and experiments is available at \url{https://github.com/alessioGalatolo/activ-pref-probe}.

\paragraph{Models.} We experiment with Llama 3, Gemma 3, Qwen 3 at varying scales from 0.6B to 14B parameters.

\paragraph{Probe training.} Unless otherwise stated, the probe is trained on 500 examples, with activations from the three layers following the middle one (e.g., for a 16-layer model, we use layers 7, 8 and 9). The probe is then used to label (up to) 50K examples.

\paragraph{Datasets.} We begin with a common SFT training using the instruction tuning dataset `Alpaca' \cite{alpaca}. This dataset is useful to provide the model with \textit{some} instruction following behaviour without focusing on the harmlessness-helpfulness that is generally injected in the preference optimisation phase. After SFT, we proceed with preference optimisation, where we experiment with HH-RLHF \cite{bai2022training}, UltraFeedback \cite{cui2023ultrafeedback}, and Nectar \cite{nectar}.

\paragraph{Methods.} We compare DPO \citep{rafailov2023direct}, IPO \citep{azar2024general}, CPO \citep{xu2024contrastive}, and KTO \citep{ethayarajh2024kto}. For all the experiments, we use LoRA \cite{hu2022lora} to reduce computational demand.

\paragraph{Evaluation.} We evaluate using LLM-as-a-judge on 500 samples from the original dataset using `Qwen 2.5 14B Instruct' \cite{qwen25} as the judge. While we also tested other judges, we found this to be a good trade-off between performance, consistency and size. Our initial tests utilised Llama 3.2 1B/3B (Instruct), however, both yielded high variability and tie rates. After testing smaller models from the Qwen 2.5 family, we selected the 14B variant as the most reliable. Nevertheless, we report high consistency even among the different judges, where most judges would, on average, select the same winner model.

\paragraph{Baselines.}
\begin{itemize}
    \item \textbf{Original labels}: Standard preference optimisation with original annotations.
    \item \textbf{Random labels}: Uniform random assignment of chosen/rejected (i.e., half the samples will have swapped labels).
    \item \textbf{SFT checkpoint:} To verify that the model actually improves compared to before training.
\end{itemize}

\subsection{Comparison to Alternative Pseudo-Labelling Strategies}
\label{sec:baselines}

Before testing training using the probe's pseudo-labels, a natural question is whether activations provide signal beyond what is already recoverable from the model's own output distribution, or from a supervised classifier trained on the same seed set. We compare against two additional pseudo-labelling baselines using the same 500-example budget: (1) \textbf{LLM-as-judge self-labelling}, where the instruction-tuned model itself re-labels pairs given a rubric, optionally optimised with GEPA \citep{gepa} over 300 rollouts; and (2) a \textbf{supervised classifier head}, obtained by fine-tuning a LoRA classifier on top of the base model using the same 500 labelled pairs.

\begin{table}[h]
\centering
\caption{LLM-as-judge self-labelling accuracy (\%) across model sizes and datasets.}
\label{tab:judge_baseline}
\adjustbox{width=\columnwidth}{
\begin{tabular}{lccc}
\toprule
Model & HH-RLHF & Nectar & UltraFeedback \\
\midrule
Qwen 3 0.6B  & 49.6 & 49.8 & 50.2 \\
Llama 3.2 1B & 48.2 & 50.0 & 50.0 \\
Llama 3.2 3B & 48.7 & 66.7 & 54.4 \\
Gemma 3 4B   & 51.2 & 85.0 & 63.4 \\
Qwen 3 14B   & 47.2 & 89.9 & 71.2 \\
\bottomrule
\end{tabular}
}
\end{table}

Table~\ref{tab:judge_baseline} shows that small judges sit at or below chance across all three datasets; only larger variants become competitive, and none exceed chance on HH-RLHF, where our probe reaches 57--62\% (close to the $\sim$63\% inter-annotator ceiling). GEPA optimisation does not change this picture: Qwen~3~0.6B stays within $\pm0.2$ pp on every dataset, and Gemma~3~4B gains at most 2 pp on UltraFeedback while losing on Nectar. This indicates the failure is not a prompting artefact: small models do not expose the preference signal through their output distribution, even though it is linearly present in their activations (Section~\ref{sec:geometry}).

\begin{table}[h]
\centering
\caption{Pseudo-labelling accuracy (\%): LoRA classifier head vs.\ our linear probe, both trained on the same 500 labels (Qwen 3 0.6B).}
\label{tab:classifier_baseline}
\adjustbox{width=\columnwidth}{
\begin{tabular}{lccc}
\toprule
Method & HH-RLHF & Nectar & UltraFeedback \\
\midrule
LoRA classifier head & 56.5 & \textbf{85.5} & 57.2 \\
Linear probe (ours) & \textbf{57.7} & 75.3 & \textbf{62.1} \\
\bottomrule
\end{tabular}
}
\end{table}

Training a classifier (Table~\ref{tab:classifier_baseline}) underperforms our linear probe in two of three settings, despite substantially longer training and orders of magnitude more trainable parameters. Together, these two baselines address the central question of this comparison: preference signal in the activations is not fully recoverable from the model's output distribution, and a comparable-budget supervised classifier does not reliably recover it either.

\subsection{Equal-Budget Comparison}
\label{sec:equal_budget}

To isolate the value of label propagation from any advantage due to dataset size, we run a head-to-head comparison in which both our method and the \textbf{Original} baseline receive exactly 500 labelled examples. The original-label baseline trains directly on those 500 pairs; our probe-labelled model uses the same 500 examples for probe training and then propagates labels to the full 50K unlabelled corpus. We use HH-RLHF and DPO, as the most commonly used dataset and PO method, respectively. Results are reported in Table~\ref{tab:equal_budget}.

\begin{table}[h]
\centering
\caption{Equal-budget comparison (500 labelled examples each). Win/Tie/Loss rates for the probe-labelled model versus a model trained directly on 500 original-labelled examples, using DPO on HH-RLHF.}
\label{tab:equal_budget}

\adjustbox{width=\columnwidth}{
\begin{tabular}{lccc}
\toprule
Model & Probe wins (\%) & Original wins (\%) & Ties (\%) \\
\midrule
Qwen 3 14B   & \textbf{47.0} & 28.6 & 24.4 \\
Gemma 3 4B   & \textbf{38.0} & 37.6 & 24.4 \\
Llama 3.2 3B & \textbf{41.2} & 30.2 & 28.5 \\
Llama 3.2 1B & \textbf{33.8} & 30.4 & 35.8 \\
Qwen 3 0.6B  & 27.6 & \textbf{30.4} & 42.0 \\
\bottomrule
\end{tabular}
}
\end{table}

The probe-labelled approach outperforms direct training on the small labelled set in 4 out of 5 model settings, with win-rate advantages of up to $+11.2\%$ (excluding ties). The only exception is Qwen 3 0.6B. We attribute this primarily to the model's size, which likely makes it more susceptible to noise and training degeneration.

\subsection{500 vs 50K Labels}
Motivated by the success of our method over this small set of experiments, we expand our investigation to a harder setting. Now, our method is still only given 500 labelled samples \textit{but} the baseline(s) are now given the full set of 50K labelled samples.

We show in Table~\ref{tab:main} and \ref{tab:sft} the results using HH-RLHF, while results on other datasets are presented in Table~\ref{tab:other_datasets}.

\begin{table*}[t]
    \centering
    \caption{Probe labelled models \textit{win+tie} rate in percentage \% (tie rate also in parentheses) compared to original-labelled models or randomly labelled model after full training on HH-RLHF. Higher indicates probe labels perform better. We highlight in \textbf{bold} the times the probe-labelled model has a higher win-rate than its baseline (i.e., excluding ties). We mark with a `*' the runs where one or both models occasionally express degenerated outputs.}
    \label{tab:main}
    \adjustbox{width=\textwidth}{
        \begin{tabular}{lccccccccc}
            \toprule
            & & \multicolumn{4}{c}{\textbf{Original}} & \multicolumn{4}{c}{\textbf{Random}}\\
            \cmidrule(lr){3-6} \cmidrule(lr){7-10}
            \textbf{Family} & \textbf{Size} & DPO & IPO & CPO & KTO & DPO & IPO & CPO & KTO \\
            \midrule
            Qwen 3 & 14B & 34.4 (2.8) & \textbf{86 (1.4)} & 52.4 (17.6) & 50.6 (12.2) & \textbf{57.4 (9.8)} & \textbf{90.6 (2.2)} & \textbf{68.8 (26.6)} & \textbf{70.6 (13.8)}\\
            Gemma 3 & 4B & 33 (9) & \textbf{94* (73)} & 45.4 (17.2) & 56.2 (25) & \textbf{67.2 (31.2)} & 42.8* (20) & \textbf{67.6 (25.6)} & \textbf{98.2 (24.4)} \\
            Llama 3.2 & 3B & 39.4 (5.8) & 55.2 (20.8) & 38.8 (12.6) & 48.2 (7.2) & \textbf{67.4 (28.2)} & 46.2 (18)  & 61 (26) & \textbf{63.8 (15.2)}\\
            Llama 3.2 & 1B & 37.3 (11.4) & \textbf{92.6 (31.8)} & 49.4 (21.4) & 48 (23) & \textbf{67.6 (29.8)} & \textbf{63.6 (19.2)} & \textbf{71.5 (30.2)} & \textbf{94 (15.6)}\\
            Qwen 3 & 0.6B & 37 (20) & 63.8* (41) & 69* (49.4) & 60* (32) & 70.8 (45) & \textbf{72.4* (29)} & \textbf{79 (55.2)} & \textbf{98.4 (44)}\\
            \bottomrule
        \end{tabular}
    }
\end{table*}

\begin{table}[t]
    \centering
    \caption{Probe labelled models \textit{win+tie} rate in percentage \% (tie rate also in parentheses) compared to Supervised Fine-Tuned (SFT) checkpoint on HH-RLHF. Higher indicates probe labels perform better. We highlight in \textbf{bold} the times the probe-labelled model has a higher win-rate than its baseline (i.e., excluding ties).}
    \label{tab:sft}
    \adjustbox{width=\columnwidth}{
        \begin{tabular}{lccccc}
            \toprule
            
            & & \multicolumn{4}{c}{SFT}\\
            \cmidrule(lr){3-6}
            Family & Size & DPO & IPO & CPO & KTO \\
            \midrule
            Qwen 3 & 14B & \textbf{80.9 (9.80)} & \textbf{91.6 (0.8)} & 52.6 (16.8) & \textbf{65 (10.4)} \\
            Gemma 3 & 4B & \textbf{69.8 (30)} & 32.4* (11) & 60.2 (21.8) & \textbf{58.6 (15)} \\
            Llama 3.2 & 3B & \textbf{70.4 (28.2)} & 50.6 (21.2) & 55.2 (20.6) & \textbf{62.4 (18.2)} \\
            Llama 3.2 & 1B & \textbf{73.8 (33.4)} & \textbf{63.8 (24.6)} & \textbf{61.6 (22.6)} & \textbf{62.8 (20.2)} \\
            Qwen 3 & 0.6B & \textbf{77.8 (40.6)} & \textbf{73 (30.6)} & \textbf{74.4 (44.6)} & \textbf{63.6 (38.6)} \\
            \bottomrule
        \end{tabular}
    }
\end{table}
\paragraph{Probe labels outperform baselines, with IPO being the most robust method.} Models trained using our labelling method consistently outperform two out of three baselines, showing improvement over the SFT checkpoint (Table~\ref{tab:sft}) and over PO models trained with random labels. At both small and big scales, IPO with probe labels is very close and sometimes outperforms models trained on the original dataset. In contrast, DPO shows the largest gap between original and probe labels, regardless of scale. The other methods, CPO and KTO, show competitiveness between probe-labelled and original-labelled variants; while the original-labelled still exhibits better performance in most cases, the margins are smaller than those observed with DPO.

\paragraph{Small models degenerate.} The smallest model tested (Qwen 3 0.6B) is prone to degenerate outputs under both label sources. Two opposite failure modes are worth noting: IPO degenerates more often with \textit{original} labels than with probe labels, while DPO degenerates more often with \textit{probe} labels than with originals. This is consistent with each method's distinct sensitivity to the structure of the label distribution.

\paragraph{Label smoothing improves competitiveness with DPO.}

Label smoothing has been proposed to improve DPO robustness \citep{mitchell2023note}. Table~\ref{tab:smoothing} compares different smoothing values. 

\begin{table}[t]
\centering
\caption{Effect of label smoothing on DPO with probe labels (Qwen3 0.6B, HH-RLHF). Win rate (\%) against original labels. We highlight in \textbf{bold} the best results for the probe.}
\label{tab:smoothing}
\begin{tabular}{lccc}
\toprule
Label Smoothing & Win & Lose & Tie \\
\midrule
0.0 & 17 & 63 & 20 \\
0.1 & 18.4 & 57.2 & 24.4 \\
0.25 & \textbf{25.4} & 40 & 34.6 \\
0.4 & 24.6 & \textbf{32.8} & \textbf{42.6} \\
\bottomrule
\end{tabular}
\end{table}

Higher smoothing substantially improves robustness, increasing the win-tie rate from 37\% to 67.2\%. This aligns with the interpretation that DPO overfits to noisy labels.

\paragraph{Probe performance plateaus after 500 samples.}

\begin{figure}[h]
    \centering
    \includegraphics[width=\columnwidth]{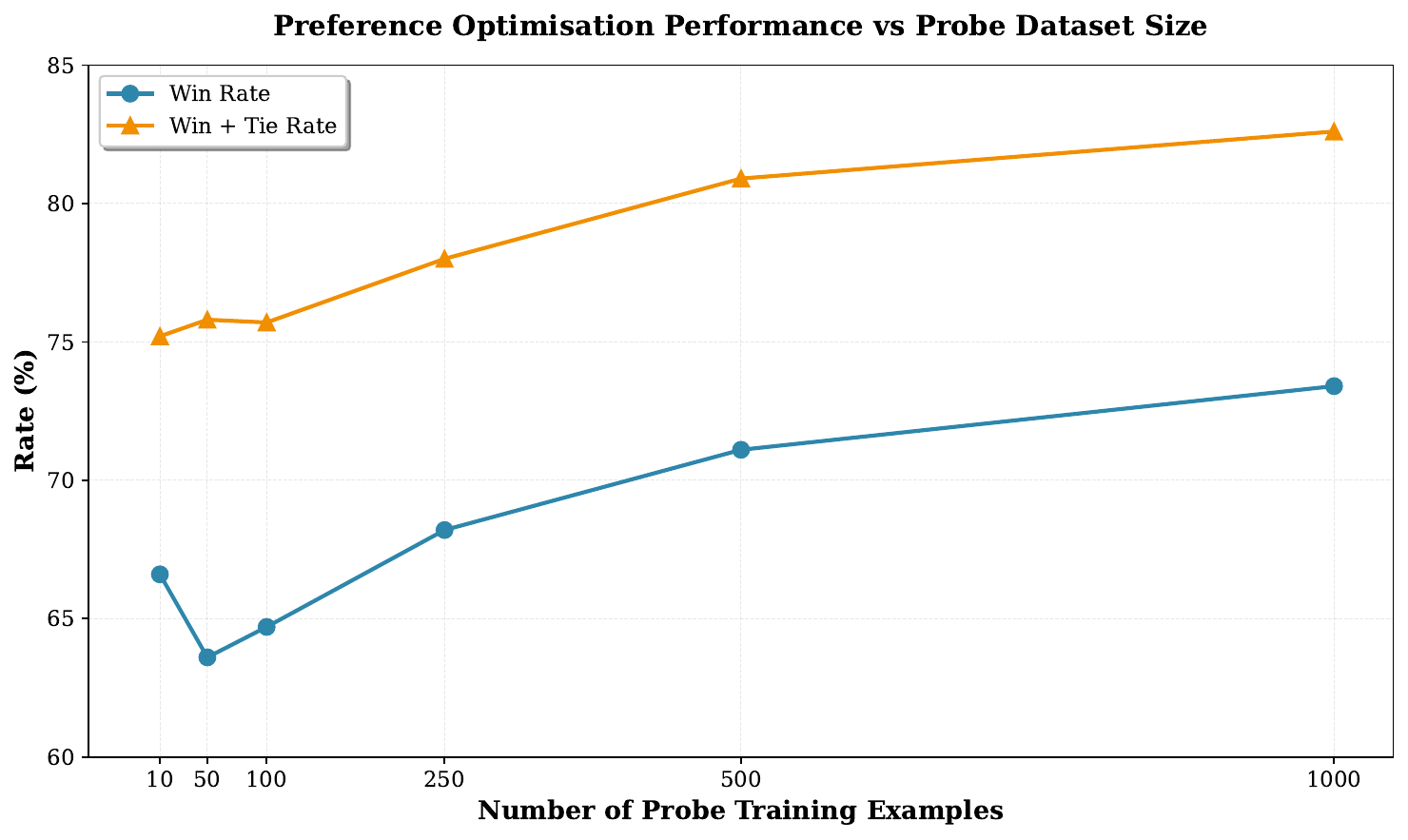}
    \caption{Effect of probe training set size on downstream preference optimisation performance. Reported is the win-rate against the SFT model. Performance starts to rise after 100 samples.}
    \label{fig:probe_size}
\end{figure}

Figure~\ref{fig:probe_size} shows a visualisation of probe performance as a function of training set size. We test the following configurations: $n \in \{10, 50, 100, 250, 500, 1000\}$. We plot the win-rate of the probe-trained model against the SFT model. While the plot shows almost monotonically increasing win-tie rate, the biggest increase in performance happens with 250 ($\sim77\%$ win-ties) training samples until 500 ($~\sim81\%$). After this boundary, increasing training size yields diminishing results (while still improving overall performance).


\paragraph{Other datasets exhibit similar performance.}
\label{sec:other_datasets}
The results on UltraFeedback and Nectar (Table~\ref{tab:other_datasets}) are very similar to those from HH-RLHF: models trained on probe labels often outperform randomly labelled datasets and, only occasionally, the models trained with the original dataset.

\begin{table*}[htbp]
  \centering
  \caption{Probe labelled models \textit{win+tie} rate in percentage \% (tie rate also in parentheses) on Nectar (above) and UltraFeedback (below).}
  \label{tab:other_datasets}
  \adjustbox{width=\textwidth}{
  \begin{tabular}{lccccccccc}
    \toprule
    \multicolumn{10}{c}{\textbf{\textit{Nectar}}} \\
    \midrule
    & & \multicolumn{4}{c}{\textbf{Original}} & \multicolumn{4}{c}{\textbf{Random}}\\
    \cmidrule(lr){3-6} \cmidrule(lr){7-10}
    \textbf{Family} & \textbf{Size} & DPO & IPO & CPO & KTO & DPO & IPO & CPO & KTO \\
    \midrule
    Qwen 3 & 14B &  26.0 (2.2) & 66.8 (41.0) & 37.8 (2.6) & 46.6 (3.2) & \textbf{63.2 (7.4)} & 15.2 (7.6) & \textbf{73.4 (4.8)} & \textbf{99.0 (1.4)} \\
    Gemma 3 & 4B & 18.6 (2.2) & 37.4 (4.6) & 27.4 (4.6) & 38.4 (3.6) & \textbf{64.4 (23.8)} & 31.0 (14.2) & 52.8 (9.8) & \textbf{98.0 (7.8)} \\
    Llama & 3B &  26.0 (3.6) & \textbf{90.4 (75.6)} & 34.6 (4.2) & \textbf{54.6 (5.6)} & \textbf{75.8 (22.2)} &  14.2 (5.6) & \textbf{62.2 (9.8)} & \textbf{77.6 (3.4)} \\
    Llama 3.2 & 1B &  36.8 (9.0) & \textbf{99.2 (26.6)} & 39.2 (9.6) & 36.0 (8.0) & \textbf{61.0 (18.6)} & 30.6 (18.4) & 56.4 (18.8) & \textbf{70.4 (13.2)} \\
    Qwen 3 & 0.6B &  38.0 (19.6) & \textbf{90.6 (50.0)} & 54.8 (20.4) & 46.8 (16.0) & \textbf{72.8 (36.4)} & 49.8 (33.4) & \textbf{70.2 (22.4)} & \textbf{93.2 (16.0)} \\
    \bottomrule
    \toprule
    \multicolumn{10}{c}{\textbf{\textit{UltraFeedback}}} \\
    \midrule
    Qwen 3 & 14B & 30.2 (3.0) & 52.4 (15.8) & 38.0 (2.4) & 19.8 (1.6) & \textbf{55.4 (6.4)} & 20.8 (8.2) & \textbf{62.4 (3.6)} & \textbf{98.2 (7.4)} \\
    Gemma 3 & 4B  & 28.4 (8.2) & 23.2 (6.0) & 41.4 (4.0) & 26.0 (2.0) & \textbf{62.0 (17.4)} & 22.4 (8.4) & 51.6 (5.2) & \textbf{79.8 (5.0)} \\
    Llama & 3B  & 32.6 (8.4) & 74.0 (73.0) & 40.6 (4.2) & 30.6 (3.2) & \textbf{61.2 (12.8)} & 8.2 (6.4) & \textbf{55.2 (7.6)} & \textbf{70.2 (6.2)} \\
    Llama 3.2 & 1B  & 39.0 (15.2) & \textbf{89.8 (34.0)} & 51.2 (17.0) & 30.6 (6.8) & 47.6 (14.6) & 19.8 (14.6) & \textbf{60.6 (18.2)} & \textbf{59.8 (13.6)} \\
    Qwen 3 & 0.6B  & 48.0 (24.2) & \textbf{75.8 (18.6)} & 47.2 (15.6) & 30.4 (7.2) & \textbf{68.6 (32.4)} & 54.4 (25.4) & \textbf{63.2 (25.0)} & \textbf{73.8 (24.0)} \\
    \bottomrule
  \end{tabular}
  }
\end{table*}

\paragraph{Where does the crossover lie?} The equal-budget (Section~\ref{sec:equal_budget}) and 500-vs-50K comparisons above represent two extremes. To locate the crossover point, we fix the probe's training budget at 500 and vary the \textit{original}-label baseline's budget over $\{500, 1\text{k}, 2.5\text{k}, 5\text{k}\}$, using DPO on HH-RLHF.

\begin{table}[t]
\centering
\caption{Win+tie (tie) rates (\%) for the probe-labelled model (fixed at 500 labels) against an original-label baseline with increasing budget, using DPO on HH-RLHF. Bold indicates the probe's strict wins.}
\label{tab:crossover}
\adjustbox{width=\columnwidth}{
\begin{tabular}{llcccc}
\toprule
Family & Size & 500 & 1K & 2.5K & 5K \\
\midrule
Qwen 3 & 14B & \textbf{71.4} (24.4) & \textbf{58.6} (11.2) & 43.6 (8.2) & 42.6 (6.4) \\
Gemma 3 & 4B  & \textbf{62.4} (24.4) & 60.8 (28.0) & 43.8 (14.0) & 34.4 (8.2) \\
Llama 3.2 & 3B & \textbf{69.7} (28.5) & \textbf{66.2} (26.4) & 49.0 (15.6) & 37.7 (11.0) \\
Llama 3.2 & 1B & \textbf{69.6} (35.8) & \textbf{69.6} (33.2) & 57.2 (27.4) & 41.0 (14.0) \\
Qwen 3 & 0.6B & 69.6 (42.0) & \textbf{72.8} (43.4) & 52.6 (37.4) & 40.2 (27.2) \\
\bottomrule
\end{tabular}
}
\end{table}

The crossover lies consistently between 1k and 2.5k gold labels across model scales. Since DPO represent our least noise-robust method, this crossover should be read as a conservative lower bound; other (dataset, method) combinations are likely to favour the probe at even higher gold-label budgets.

\subsection{Additional validation}
We provide two further validation experiments in Appendix \ref{app:additional_validation}. First, a small-scale human evaluation on HH-RLHF (250 response pairs, four annotators, three model--method configurations) confirms that the directional pattern observed in our automated evaluation holds under human judgement: IPO with probe labels outperforms both baselines, while DPO with probe labels trails the 50K original-label baseline but clearly surpasses random labelling. Second, a downstream preference optimisation experiment on the PRISM dataset assesses transfer to a culturally specific, low-resource setting, using embedding-based similarity to ground-truth completions as an evaluation proxy. The probe-trained model matches or exceeds a fully-supervised baseline in the majority of configurations, with IPO showing up to $29\%$ relative improvement despite using only 500 labelled examples. Together, these results corroborate our central finding that probe-generated labels carry genuine preference signal across diverse evaluation regimes.

\subsection{Computational Overhead}
A key practical advantage of our approach is its efficiency: probe training requires $n$ forward passes to extract activations (where $n \leq 500$) followed by fitting a linear classifier, completing in under a minute on a single GPU. Label propagation (on 50K+ samples) in our current implementation adds approximately 70\% overhead to preference optimisation training time, as we perform labelling as a separate preprocessing step for simplicity of implementation. However, this overhead can be eliminated entirely by fusing label propagation with preference optimisation. Specifically, during each training iteration, one can perform a forward pass on an unlabelled pair $(y^a, y^b)$, extract activations to obtain probe predictions, assign labels accordingly, and then compute the preference optimisation loss using the same forward pass outputs before backpropagating. This fused approach adds only the cost of a single linear probe evaluation per batch, negligible compared to the LLM forward and backward passes. We provide a sketch of this algorithm in Appendix~\ref{app:algorithm}.

\section{Conclusion}

We showed that language models encode preference information as geometrically separable clusters in their activations, that this geometry is reshaped, and in the case of non-canonical preferences, \textit{erased} by alignment. Exploiting this structure, we proposed a method for label-efficient preference adaptation: train a linear probe on a few hundred labelled examples, propagate labels to a large unlabelled corpus, and run preference optimisation. Our systematic evaluation revealed substantial differences in method robustness: IPO tolerates noisy labels while DPO degrades unless label smoothing is used. Nevertheless, training using probe labels has shown consistent improvement over SFT models and random labelling. This work enables preference adaptation with 100$\times$ fewer annotations, making personalised alignment more accessible, especially in low-resource settings.

\subsection{Limitations}
Our work has several limitations that suggest directions for future research. First, our evaluation almost entirely relies on LLM-as-a-judge, which creates tension with our motivation that LLM judges encode their own preference biases. While our small-scale human evaluation and consistency across multiple judge models provide corroborating evidence, human evaluation at the full scale of Table~\ref{tab:main} would further strengthen our claims. Second, although we demonstrate that PRISM exhibits similar activation geometry to standard datasets (Section~\ref{sec:culturally_divers}) and provide initial downstream results, human evaluation on culturally diverse populations remains infeasible without access to the original annotator communities. Finally, while our empirical findings are consistent with the linear representation hypothesis \citep{jiang2024origins, elhage2022superposition} and probing theory \citep{pimentel2020information}, providing formal convergence guarantees for preference optimisation under probe-generated noise remains an open problem that we leave for future work.

\section*{Acknowledgements}
The computations and data handling were enabled by resources provided by the National Academic Infrastructure for Supercomputing in Sweden (NAISS) including Arrhenius and Alvis, C3SE (Chalmers) partially funded by the Swedish Research Council through grant agreement no. 2022-06725.



\bibliography{references}

@article{ouyang2022training,
  title={Training language models to follow instructions with human feedback},
  author={Ouyang, Long and Wu, Jeffrey and Jiang, Xu and Almeida, Diogo and Wainwright, Carroll and Mishkin, Pamela and Zhang, Chong and Agarwal, Sandhini and Slama, Katarina and Ray, Alex and others},
  journal={Advances in neural information processing systems},
  volume={35},
  pages={27730--27744},
  year={2022}
}

@article{rafailov2023direct,
  title={Direct preference optimization: Your language model is secretly a reward model},
  author={Rafailov, Rafael and Sharma, Archit and Mitchell, Eric and Manning, Christopher D and Ermon, Stefano and Finn, Chelsea},
  journal={Advances in neural information processing systems},
  volume={36},
  pages={53728--53741},
  year={2023}
}

@inproceedings{azar2024general,
  title = 	 {A General Theoretical Paradigm to Understand Learning from Human Preferences},
  author =       {Gheshlaghi Azar, Mohammad and Daniel Guo, Zhaohan and Piot, Bilal and Munos, Remi and Rowland, Mark and Valko, Michal and Calandriello, Daniele},
  booktitle = 	 {Proceedings of The 27th International Conference on Artificial Intelligence and Statistics},
  pages = 	 {4447--4455},
  year = 	 {2024},
  editor = 	 {Dasgupta, Sanjoy and Mandt, Stephan and Li, Yingzhen},
  volume = 	 {238},
  series = 	 {Proceedings of Machine Learning Research},
  month = 	 {02--04 May},
  publisher =    {PMLR},
  url = 	 {https://proceedings.mlr.press/v238/gheshlaghi-azar24a.html}
}

@inproceedings{kirk2024prism,
 author = {Kirk, Hannah Rose and Whitefield, Alexander and R\"{o}ttger, Paul and Bean, Andrew and Margatina, Katerina and Ciro, Juan and Mosquera, Rafael and Bartolo, Max and Williams, Adina and He, He and Vidgen, Bertie and Hale, Scott A.},
 booktitle = {Advances in Neural Information Processing Systems},
 doi = {10.52202/079017-3342},
 editor = {A. Globerson and L. Mackey and D. Belgrave and A. Fan and U. Paquet and J. Tomczak and C. Zhang},
 pages = {105236--105344},
 publisher = {Curran Associates, Inc.},
 title = {The PRISM Alignment Dataset: What Participatory, Representative and Individualised Human Feedback Reveals About the Subjective and Multicultural Alignment of Large Language Models},
 volume = {37},
 year = {2024}
}

@inproceedings{sorensen2024value,
  title={Value kaleidoscope: Engaging ai with pluralistic human values, rights, and duties},
  author={Sorensen, Taylor and Jiang, Liwei and Hwang, Jena D and Levine, Sydney and Pyatkin, Valentina and West, Peter and Dziri, Nouha and Lu, Ximing and Rao, Kavel and Bhagavatula, Chandra and others},
  booktitle={Proceedings of the AAAI Conference on Artificial Intelligence},
  volume={38},
  number={18},
  pages={19937--19947},
  year={2024}
}

@article{zheng2023judging,
  title={Judging llm-as-a-judge with mt-bench and chatbot arena},
  author={Zheng, Lianmin and Chiang, Wei-Lin and Sheng, Ying and Zhuang, Siyuan and Wu, Zhanghao and Zhuang, Yonghao and Lin, Zi and Li, Zhuohan and Li, Dacheng and Xing, Eric and others},
  journal={Advances in neural information processing systems},
  volume={36},
  pages={46595--46623},
  year={2023}
}

@article{christiano2017deep,
  title={Deep reinforcement learning from human preferences},
  author={Christiano, Paul F and Leike, Jan and Brown, Tom and Martic, Miljan and Legg, Shane and Amodei, Dario},
  journal={Advances in neural information processing systems},
  volume={30},
  year={2017}
}

@inproceedings{ethayarajh2024kto,
author = {Ethayarajh, Kawin and Xu, Winnie and Muennighoff, Niklas and Jurafsky, Dan and Kiela, Douwe},
title = {Model alignment as prospect theoretic optimization},
year = {2024},
publisher = {JMLR.org},
booktitle = {Proceedings of the 41st International Conference on Machine Learning},
articleno = {504},
numpages = {18},
location = {Vienna, Austria},
series = {ICML'24}
}

@inproceedings{xu2024contrastive,
  title={Contrastive preference optimization: pushing the boundaries of LLM performance in machine translation},
  author={Xu, Haoran and Sharaf, Amr and Chen, Yunmo and Tan, Weiting and Shen, Lingfeng and Van Durme, Benjamin and Murray, Kenton and Kim, Young Jin},
  booktitle={Proceedings of the 41st International Conference on Machine Learning},
  pages={55204--55224},
  year={2024}
}

@article{bai2022constitutional,
  title={Constitutional ai: Harmlessness from ai feedback},
  author={Bai, Yuntao and Kadavath, Saurav and Kundu, Sandipan and Askell, Amanda and Kernion, Jackson and Jones, Andy and Chen, Anna and Goldie, Anna and Mirhoseini, Azalia and McKinnon, Cameron and others},
  journal={arXiv preprint arXiv:2212.08073},
  year={2022}
}

@inproceedings{lee2024rlaif,
  title={RLAIF vs. RLHF: scaling reinforcement learning from human feedback with AI feedback},
  author={Lee, Harrison and Phatale, Samrat and Mansoor, Hassan and Mesnard, Thomas and Ferret, Johan and Lu, Kellie and Bishop, Colton and Hall, Ethan and Carbune, Victor and Rastogi, Abhinav and others},
  booktitle={Proceedings of the 41st International Conference on Machine Learning},
  pages={26874--26901},
  year={2024}
}

@article{askell2021general,
  title={A general language assistant as a laboratory for alignment},
  author={Askell, Amanda and Bai, Yuntao and Chen, Anna and Drain, Dawn and Ganguli, Deep and Henighan, Tom and Jones, Andy and Joseph, Nicholas and Mann, Ben and DasSarma, Nova and others},
  journal={arXiv preprint arXiv:2112.00861},
  year={2021}
}

@inproceedings{burns2024weak,
  title={Weak-to-strong generalization: eliciting strong capabilities with weak supervision},
  author={Burns, Collin and Izmailov, Pavel and Kirchner, Jan Hendrik and Baker, Bowen and Gao, Leo and Aschenbrenner, Leopold and Chen, Yining and Ecoffet, Adrien and Joglekar, Manas and Leike, Jan and others},
  booktitle={Proceedings of the 41st International Conference on Machine Learning},
  pages={4971--5012},
  year={2024}
}

@misc{mitchell2023note,
  title={A note on dpo with noisy preferences \& relationship to ipo},
  author={Mitchell, Eric},
  year={2023},
  url={https://ericmitchell.ai/cdpo.pdf}
}

@inproceedings{hewitt2019structural,
  title={A structural probe for finding syntax in word representations},
  author={Hewitt, John and Manning, Christopher D},
  booktitle={Proceedings of the 2019 Conference of the North American Chapter of the Association for Computational Linguistics: Human Language Technologies, Volume 1 (Long and Short Papers)},
  pages={4129--4138},
  year={2019}
}

@inproceedings{tenney2019bert,
  title={BERT Rediscovers the Classical NLP Pipeline},
  author={Tenney, Ian and Das, Dipanjan and Pavlick, Ellie},
  booktitle={Proceedings of the 57th Annual Meeting of the Association for Computational Linguistics},
  pages={4593--4601},
  year={2019}
}

@article{meng2022locating,
  title={Locating and editing factual associations in gpt},
  author={Meng, Kevin and Bau, David and Andonian, Alex and Belinkov, Yonatan},
  journal={Advances in neural information processing systems},
  volume={35},
  pages={17359--17372},
  year={2022}
}

@inproceedings{azaria2023internal,
  title={The Internal State of an LLM Knows When It’s Lying},
  author={Azaria, Amos and Mitchell, Tom},
  booktitle={Findings of the Association for Computational Linguistics: EMNLP 2023},
  pages={967--976},
  year={2023}
}

@inproceedings{
marks2024geometry,
title={The Geometry of Truth: Emergent Linear Structure in Large Language Model Representations of True/False Datasets},
author={Samuel Marks and Max Tegmark},
booktitle={First Conference on Language Modeling},
year={2024},
url={https://openreview.net/forum?id=aajyHYjjsk}
}

@article{arditi2024refusal,
  title={Refusal in language models is mediated by a single direction},
  author={Arditi, Andy and Obeso, Oscar and Syed, Aaquib and Paleka, Daniel and Panickssery, Nina and Gurnee, Wes and Nanda, Neel},
  journal={Advances in Neural Information Processing Systems},
  volume={37},
  pages={136037--136083},
  year={2024}
}

@article{bai2022training,
  title={Training a helpful and harmless assistant with reinforcement learning from human feedback},
  author={Bai, Yuntao and Jones, Andy and Ndousse, Kamal and Askell, Amanda and Chen, Anna and DasSarma, Nova and Drain, Dawn and Fort, Stanislav and Ganguli, Deep and Henighan, Tom and others},
  journal={arXiv preprint arXiv:2204.05862},
  year={2022}
}

@misc{cui2023ultrafeedback,
      title={UltraFeedback: Boosting Language Models with High-quality Feedback}, 
      author={Ganqu Cui and Lifan Yuan and Ning Ding and Guanming Yao and Wei Zhu and Yuan Ni and Guotong Xie and Zhiyuan Liu and Maosong Sun},
      year={2023},
      eprint={2310.01377},
      archivePrefix={arXiv},
      primaryClass={cs.CL}
}

@article{llama3,
  title={The llama 3 herd of models},
  author={{Llama Team, AI @ Meta}},
  journal={arXiv preprint arXiv:2407.21783},
  year={2024}
}

@misc{qwen25,
      title={Qwen2.5 Technical Report}, 
      author={{Qwen Team}},
      year={2025},
      eprint={2412.15115},
      archivePrefix={arXiv},
      primaryClass={cs.CL},
      url={https://arxiv.org/abs/2412.15115}, 
}

@misc{qwen3,
      title={Qwen3 Technical Report}, 
      author={{Qwen Team}},
      year={2025},
      eprint={2505.09388},
      archivePrefix={arXiv},
      primaryClass={cs.CL},
      url={https://arxiv.org/abs/2505.09388}, 
}

@article{van2008visualizing,
  author  = {Laurens van der Maaten and Geoffrey Hinton},
  title   = {Visualizing Data using t-SNE},
  journal = {Journal of Machine Learning Research},
  year    = {2008},
  volume  = {9},
  number  = {86},
  pages   = {2579--2605},
  url     = {http://jmlr.org/papers/v9/vandermaaten08a.html}
}

@inproceedings{
hu2022lora,
title={Lo{RA}: Low-Rank Adaptation of Large Language Models},
author={Edward J Hu and yelong shen and Phillip Wallis and Zeyuan Allen-Zhu and Yuanzhi Li and Shean Wang and Lu Wang and Weizhu Chen},
booktitle={International Conference on Learning Representations},
year={2022},
url={https://openreview.net/forum?id=nZeVKeeFYf9}
}

@misc{alpaca,
  author = {Rohan Taori and Ishaan Gulrajani and Tianyi Zhang and Yann Dubois and Xuechen Li and Carlos Guestrin and Percy Liang and Tatsunori B. Hashimoto },
  title = {Stanford Alpaca: An Instruction-following LLaMA model},
  year = {2023},
  publisher = {GitHub},
  journal = {GitHub repository},
  howpublished = {\url{https://github.com/tatsu-lab/stanford_alpaca}},
}

@article{pca,
  title={LIII. On lines and planes of closest fit to systems of points in space},
  author={Pearson, Karl},
  journal={The London, Edinburgh, and Dublin philosophical magazine and journal of science},
  volume={2},
  number={11},
  pages={559--572},
  year={1901},
  publisher={Taylor \& Francis}
}

@inproceedings{
gurnee2024language,
title={Language Models Represent Space and Time},
author={Wes Gurnee and Max Tegmark},
booktitle={The Twelfth International Conference on Learning Representations},
year={2024},
url={https://openreview.net/forum?id=jE8xbmvFin}
}

@misc{wang2025response,
      title={Response Uncertainty and Probe Modeling: Two Sides of the Same Coin in LLM Interpretability?}, 
      author={Yongjie Wang and Yibo Wang and Xin Zhou and Zhiqi Shen},
      year={2025},
      eprint={2505.18575},
      archivePrefix={arXiv},
      primaryClass={cs.AI},
      url={https://arxiv.org/abs/2505.18575}, 
}

@inproceedings{
dakhmouche2025can,
title={Can Linear Probes Measure {LLM} Uncertainty ?},
author={Ramzi Dakhmouche and Adrien Letellier and Hossein Gorji},
booktitle={NeurIPS 2025 Workshop MLxOR: Mathematical Foundations and Operational Integration of Machine Learning for Uncertainty-Aware Decision-Making},
year={2025},
url={https://openreview.net/forum?id=xrRIwEQJ5D}
}

@misc{cencerrado2025answerneededpredictingllm,
      title={No Answer Needed: Predicting LLM Answer Accuracy from Question-Only Linear Probes}, 
      author={Iván Vicente Moreno Cencerrado and Arnau Padrés Masdemont and Anton Gonzalvez Hawthorne and David Demitri Africa and Lorenzo Pacchiardi},
      year={2025},
      eprint={2509.10625},
      archivePrefix={arXiv},
      primaryClass={cs.CL},
      url={https://arxiv.org/abs/2509.10625}, 
}

@inproceedings{li-etal-2025-generation,
    title = "From Generation to Judgment: Opportunities and Challenges of {LLM}-as-a-judge",
    author = "Li, Dawei  and
      Jiang, Bohan  and
      Huang, Liangjie  and
      Beigi, Alimohammad  and
      Zhao, Chengshuai  and
      Tan, Zhen  and
      Bhattacharjee, Amrita  and
      Jiang, Yuxuan  and
      Chen, Canyu  and
      Wu, Tianhao  and
      Shu, Kai  and
      Cheng, Lu  and
      Liu, Huan",
    editor = "Christodoulopoulos, Christos  and
      Chakraborty, Tanmoy  and
      Rose, Carolyn  and
      Peng, Violet",
    booktitle = "Proceedings of the 2025 Conference on Empirical Methods in Natural Language Processing",
    month = nov,
    year = "2025",
    address = "Suzhou, China",
    publisher = "Association for Computational Linguistics",
    url = "https://aclanthology.org/2025.emnlp-main.138/",
    doi = "10.18653/v1/2025.emnlp-main.138",
    pages = "2757--2791",
    ISBN = "979-8-89176-332-6"
}

@inproceedings{
nectar,
title={Starling-7B: Improving Helpfulness and Harmlessness with {RLAIF}},
author={Banghua Zhu and Evan Frick and Tianhao Wu and Hanlin Zhu and Karthik Ganesan and Wei-Lin Chiang and Jian Zhang and Jiantao Jiao},
booktitle={First Conference on Language Modeling},
year={2024},
url={https://openreview.net/forum?id=GqDntYTTbk}
}

@misc{gemma3,
      title={Gemma 3 Technical Report}, 
      author={{Gemma Team}},
      year={2025},
      eprint={2503.19786},
      archivePrefix={arXiv},
      primaryClass={cs.CL},
      url={https://arxiv.org/abs/2503.19786}, 
}

@article{tversky1992advances,
  title={Advances in prospect theory: Cumulative representation of uncertainty},
  author={Tversky, Amos and Kahneman, Daniel},
  journal={Journal of Risk and uncertainty},
  volume={5},
  number={4},
  pages={297--323},
  year={1992},
  publisher={Springer}
}

@inproceedings{hollinsworth-etal-2024-language,
    title = "Language Models Linearly Represent Sentiment",
    author = "Tigges, Curt  and
      Hollinsworth, Oskar J.  and
      Geiger, Atticus  and
      Nanda, Neel",
    editor = "Belinkov, Yonatan  and
      Kim, Najoung  and
      Jumelet, Jaap  and
      Mohebbi, Hosein  and
      Mueller, Aaron  and
      Chen, Hanjie",
    booktitle = "Proceedings of the 7th BlackboxNLP Workshop: Analyzing and Interpreting Neural Networks for NLP",
    month = nov,
    year = "2024",
    address = "Miami, Florida, US",
    publisher = "Association for Computational Linguistics",
    url = "https://aclanthology.org/2024.blackboxnlp-1.5/",
    doi = "10.18653/v1/2024.blackboxnlp-1.5",
    pages = "58--87"
}

@article{
fisch2025robust,
title={Robust Preference Optimization through Reward Model Distillation},
author={Adam Fisch and Jacob Eisenstein and Vicky Zayats and Alekh Agarwal and Ahmad Beirami and Chirag Nagpal and Peter Shaw and Jonathan Berant},
journal={Transactions on Machine Learning Research},
issn={2835-8856},
year={2025},
url={https://openreview.net/forum?id=E2zKNuwNDc},
note={}
}

@inproceedings{maiya-etal-2025-improving,
    title = "Improving Preference Extraction In {LLM}s By Identifying Latent Knowledge Through Classifying Probes",
    author = "Maiya, Sharan  and
      Liu, Yinhong  and
      Debnath, Ramit  and
      Korhonen, Anna",
    editor = "Che, Wanxiang  and
      Nabende, Joyce  and
      Shutova, Ekaterina  and
      Pilehvar, Mohammad Taher",
    booktitle = "Proceedings of the 63rd Annual Meeting of the Association for Computational Linguistics (Volume 1: Long Papers)",
    month = jul,
    year = "2025",
    address = "Vienna, Austria",
    publisher = "Association for Computational Linguistics",
    url = "https://aclanthology.org/2025.acl-long.444/",
    doi = "10.18653/v1/2025.acl-long.444",
    pages = "9061--9081",
    ISBN = "979-8-89176-251-0"
}

@inproceedings{jiang2024origins,
  title={On the Origins of Linear Representations in Large Language Models},
  author={Jiang, Yibo and Rajendran, Goutham and Ravikumar, Pradeep Kumar and Aragam, Bryon and Veitch, Victor},
  booktitle={International Conference on Machine Learning},
  pages={21879--21911},
  year={2024},
  organization={PMLR}
}

@article{elhage2022superposition,
  title={Toy models of superposition},
  author={Elhage, Nelson and Hume, Tristan and Olsson, Catherine and Schiefer, Nicholas and Henighan, Tom and Kravec, Shauna and Hatfield-Dodds, Zac and Lasenby, Robert and Drain, Dawn and Chen, Carol and others},
  journal={arXiv preprint arXiv:2209.10652},
  year={2022}
}

@inproceedings{pimentel2020information,
    title = "Information-Theoretic Probing for Linguistic Structure",
    author = "Pimentel, Tiago  and
      Valvoda, Josef  and
      Maudslay, Rowan Hall  and
      Zmigrod, Ran  and
      Williams, Adina  and
      Cotterell, Ryan",
    editor = "Jurafsky, Dan  and
      Chai, Joyce  and
      Schluter, Natalie  and
      Tetreault, Joel",
    booktitle = "Proceedings of the 58th Annual Meeting of the Association for Computational Linguistics",
    month = jul,
    year = "2020",
    address = "Online",
    publisher = "Association for Computational Linguistics",
    url = "https://aclanthology.org/2020.acl-main.420/",
    doi = "10.18653/v1/2020.acl-main.420",
    pages = "4609--4622"
}

@inproceedings{oasst2,
 author = {K\"{o}pf, Andreas and Kilcher, Yannic and von R\"{u}tte, Dimitri and Anagnostidis, Sotiris and Tam, Zhi Rui and Stevens, Keith and Barhoum, Abdullah and Nguyen, Duc and Stanley, Oliver and Nagyfi, Rich\'{a}rd and ES, Shahul and Suri, Sameer and Glushkov, David and Dantuluri, Arnav and Maguire, Andrew and Schuhmann, Christoph and Nguyen, Huu and Mattick, Alexander},
 booktitle = {Advances in Neural Information Processing Systems},
 editor = {A. Oh and T. Naumann and A. Globerson and K. Saenko and M. Hardt and S. Levine},
 pages = {47669--47681},
 publisher = {Curran Associates, Inc.},
 title = {OpenAssistant Conversations - Democratizing Large Language Model Alignment},
 url = {https://proceedings.neurips.cc/paper_files/paper/2023/file/949f0f8f32267d297c2d4e3ee10a2e7e-Paper-Datasets_and_Benchmarks.pdf},
 volume = {36},
 year = {2023}
}

@inproceedings{muhammad-etal-2023-afrisenti,
    title = "{A}fri{S}enti: A {T}witter Sentiment Analysis Benchmark for {A}frican Languages",
    author = "Muhammad, Shamsuddeen Hassan  and
      Abdulmumin, Idris  and
      Ayele, Abinew Ali  and
      Ousidhoum, Nedjma  and
      Adelani, David Ifeoluwa  and
      Yimam, Seid Muhie  and
      Ahmad, Ibrahim Sa'id  and
      Beloucif, Meriem  and
      Mohammad, Saif M.  and
      Ruder, Sebastian  and
      Hourrane, Oumaima  and
      Brazdil, Pavel  and
      Jorge, Alipio  and
      Ali, Felermino D{\'a}rio M{\'a}rio Ant{\'o}nio  and
      David, Davis  and
      Osei, Salomey  and
      Shehu Bello, Bello  and
      Ibrahim, Falalu  and
      Gwadabe, Tajuddeen  and
      Rutunda, Samuel  and
      Belay, Tadesse  and
      Messelle, Wendimu Baye  and
      Balcha, Hailu Beshada  and
      Chala, Sisay Adugna  and
      Gebremichael, Hagos Tesfahun  and
      Opoku, Bernard  and
      Arthur, Stephen",
    editor = "Bouamor, Houda  and
      Pino, Juan  and
      Bali, Kalika",
    booktitle = "Proceedings of the 2023 Conference on Empirical Methods in Natural Language Processing",
    month = dec,
    year = "2023",
    address = "Singapore",
    publisher = "Association for Computational Linguistics",
    url = "https://aclanthology.org/2023.emnlp-main.862/",
    doi = "10.18653/v1/2023.emnlp-main.862",
    pages = "13968--13981"
}

@misc{llama2,
      title={Llama 2: Open Foundation and Fine-Tuned Chat Models}, 
      author={Hugo Touvron and Louis Martin and Kevin Stone and Peter Albert and Amjad Almahairi and Yasmine Babaei and Nikolay Bashlykov and Soumya Batra and Prajjwal Bhargava and Shruti Bhosale and Dan Bikel and Lukas Blecher and Cristian Canton Ferrer and Moya Chen and Guillem Cucurull and David Esiobu and Jude Fernandes and Jeremy Fu and Wenyin Fu and Brian Fuller and Cynthia Gao and Vedanuj Goswami and Naman Goyal and Anthony Hartshorn and Saghar Hosseini and Rui Hou and Hakan Inan and Marcin Kardas and Viktor Kerkez and Madian Khabsa and Isabel Kloumann and Artem Korenev and Punit Singh Koura and Marie-Anne Lachaux and Thibaut Lavril and Jenya Lee and Diana Liskovich and Yinghai Lu and Yuning Mao and Xavier Martinet and Todor Mihaylov and Pushkar Mishra and Igor Molybog and Yixin Nie and Andrew Poulton and Jeremy Reizenstein and Rashi Rungta and Kalyan Saladi and Alan Schelten and Ruan Silva and Eric Michael Smith and Ranjan Subramanian and Xiaoqing Ellen Tan and Binh Tang and Ross Taylor and Adina Williams and Jian Xiang Kuan and Puxin Xu and Zheng Yan and Iliyan Zarov and Yuchen Zhang and Angela Fan and Melanie Kambadur and Sharan Narang and Aurelien Rodriguez and Robert Stojnic and Sergey Edunov and Thomas Scialom},
      year={2023},
      eprint={2307.09288},
      archivePrefix={arXiv},
      primaryClass={cs.CL},
      url={https://arxiv.org/abs/2307.09288}, 
}

@inproceedings{reimers2019sentencebert,
    title = "Sentence-{BERT}: Sentence Embeddings using {S}iamese {BERT}-Networks",
    author = "Reimers, Nils  and
      Gurevych, Iryna",
    editor = "Inui, Kentaro  and
      Jiang, Jing  and
      Ng, Vincent  and
      Wan, Xiaojun",
    booktitle = "Proceedings of the 2019 Conference on Empirical Methods in Natural Language Processing and the 9th International Joint Conference on Natural Language Processing (EMNLP-IJCNLP)",
    month = nov,
    year = "2019",
    address = "Hong Kong, China",
    publisher = "Association for Computational Linguistics",
    url = "https://aclanthology.org/D19-1410/",
    doi = "10.18653/v1/D19-1410",
    pages = "3982--3992"
}

@inproceedings{gepa,
title={{GEPA}: Reflective Prompt Evolution Can Outperform Reinforcement Learning},
author={Lakshya A Agrawal and Shangyin Tan and Dilara Soylu and Noah Ziems and Rishi Khare and Krista Opsahl-Ong and Arnav Singhvi and Herumb Shandilya and Michael J Ryan and Meng Jiang and Christopher Potts and Koushik Sen and Alex Dimakis and Ion Stoica and Dan Klein and Matei Zaharia and Omar Khattab},
booktitle={The Fourteenth International Conference on Learning Representations},
year={2026},
url={https://openreview.net/forum?id=RQm2KQTM5r}
}

\clearpage
\appendix
\clearpage
\appendix

\section{Additional Results on Activation Geometry}
\label{app:additional_results}

\subsection{Sanity Check on Clustering}
\label{app:sanity_check}
\begin{figure*}[t]
    \centering
    \includegraphics[width=1\linewidth]{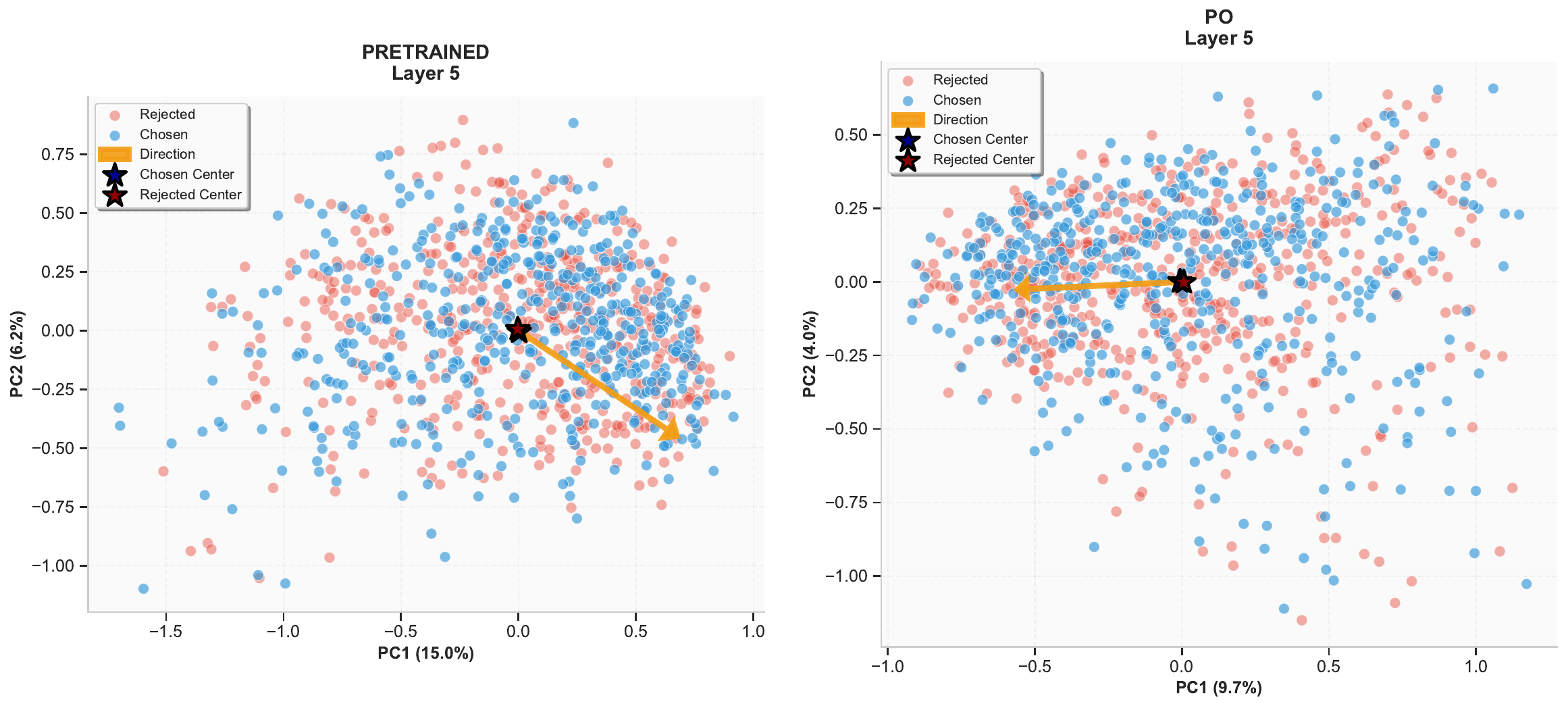}
    \caption{\textbf{Sanity check:} example clustering with labels randomly swapped. The indistinguishable nature of the clusters and centroids confirms that the separation observed in Figure~\ref{fig:clusters} arises from the contrastive structure of preferred and dis-preferred responses, and is not an artefact of unrelated activation statistics.}
    \label{fig:placeholder}
\end{figure*}

To verify that the activation separation observed in Figure~\ref{fig:clusters} is driven by genuine preference signal rather than incidental distributional differences, we repeat the t-SNE and PCA visualisations with the chosen/rejected labels randomly swapped. As shown in Figure~\ref{fig:placeholder}, the two clusters become completely indistinguishable and their centroids coincide. This confirms that the structured geometry reported in Section~\ref{sec:geometry} is a direct consequence of the contrastive nature of positive and negative examples, not an artefact of unrelated activation statistics.

\subsection{Layer-wise Probe Accuracy}
\label{app:layerwise}

\begin{figure}[t]
    \centering
    \includegraphics[width=\columnwidth]{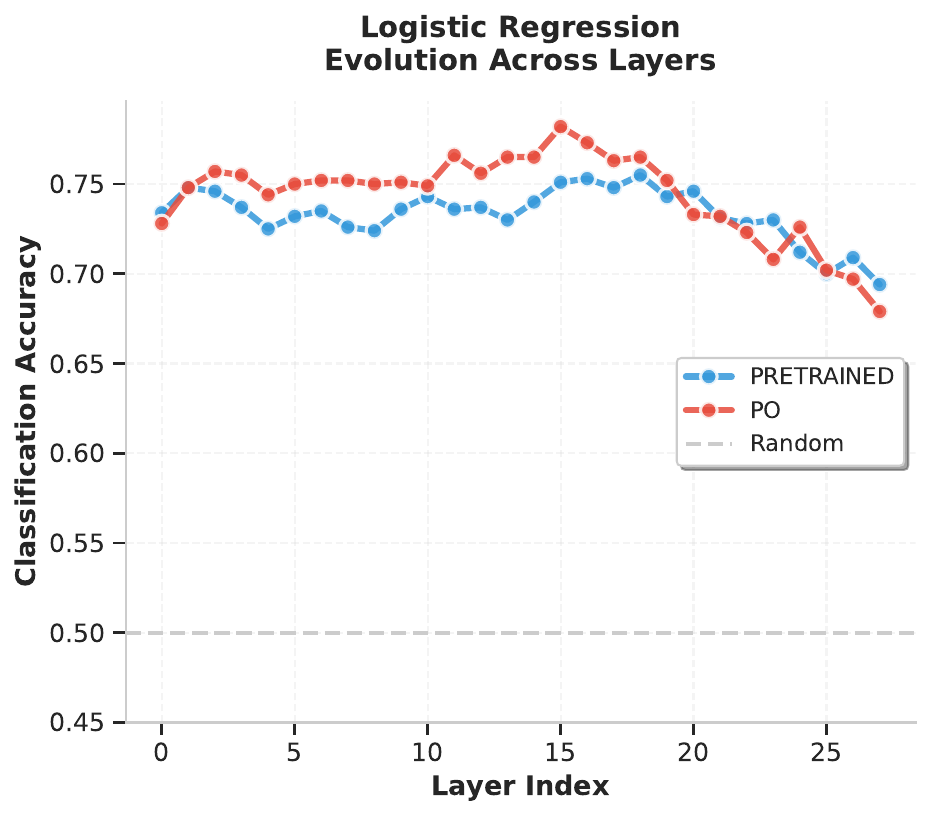}
    \caption{Linear probe accuracy across layers on Qwen~3~0.6B (Nectar dataset) for the pretrained and PO checkpoints. Accuracy peaks around layer~15, with the PO model showing higher accuracy consistently across most layers.}
    \label{fig:layerwise}
\end{figure}

Figure~\ref{fig:layerwise} presents linear probe classification accuracy across all layers of Qwen~3~0.6B on the Nectar dataset, comparing pretrained and preference-optimised (PO) checkpoints. Accuracy rises steeply in the early layers, peaks in the middle-to-late portion of the network (around layer~15 out of~28), and then plateaus or slightly declines towards the final layers. This pattern is consistent with findings from general representation probing studies~\citep{gurnee2024language}, where task-relevant information tends to be most linearly accessible in intermediate representations. The PO model shows consistently higher probe accuracy across most layers, suggesting that preference optimisation sharpens the linear separability of chosen and rejected responses in the activation space---though, as discussed in Section~\ref{sec:geometry}, the difference between pretrained and PO models is modest compared to the effect of the dataset.

These results inform our practical recommendation to extract activations from layers $\lfloor L/2 \rfloor$ to $\lfloor 2L/3 \rfloor$ when a validation set is unavailable, as this range reliably captures near-peak probe performance across all model families and datasets we examined.

\subsection{Kernel-MMD Distributional Test}
\label{app:mmd}
Table~\ref{tab:mmd} reports the full kernel-MMD results across all nine (model, dataset) combinations, complementing the univariate PC1 test in Section~\ref{sec:geometry}.

\begin{table}[h]
\centering
\caption{Kernel-MMD two-sample test between chosen and rejected activations (RBF kernel, median-heuristic bandwidth, 200 permutations, $n=250$ per class).}
\label{tab:mmd}

\adjustbox{width=\columnwidth}{
\begin{tabular}{llcc}
\toprule
Model & Dataset & MMD$^2$ & Permutation $p$ \\
\midrule
Llama & UltraFeedback & 0.0012 & 0.110 \\
Llama & HH-RLHF & 0.0044 & 0.005 \\
Llama & Nectar & 0.0407 & 0.005 \\
Gemma & UltraFeedback & 0.0049 & 0.020 \\
Gemma & HH-RLHF & 0.0087 & 0.005 \\
Gemma & Nectar & 0.0789 & 0.005 \\
Qwen & UltraFeedback & 0.0022 & 0.050 \\
Qwen & HH-RLHF & 0.0052 & 0.005 \\
Qwen & Nectar & 0.0334 & 0.005 \\
\bottomrule
\end{tabular}
}
\end{table}

\subsection{Full Main Results Illustration}

\begin{figure*}[!t]
    \centering
    \includegraphics[width=\textwidth]{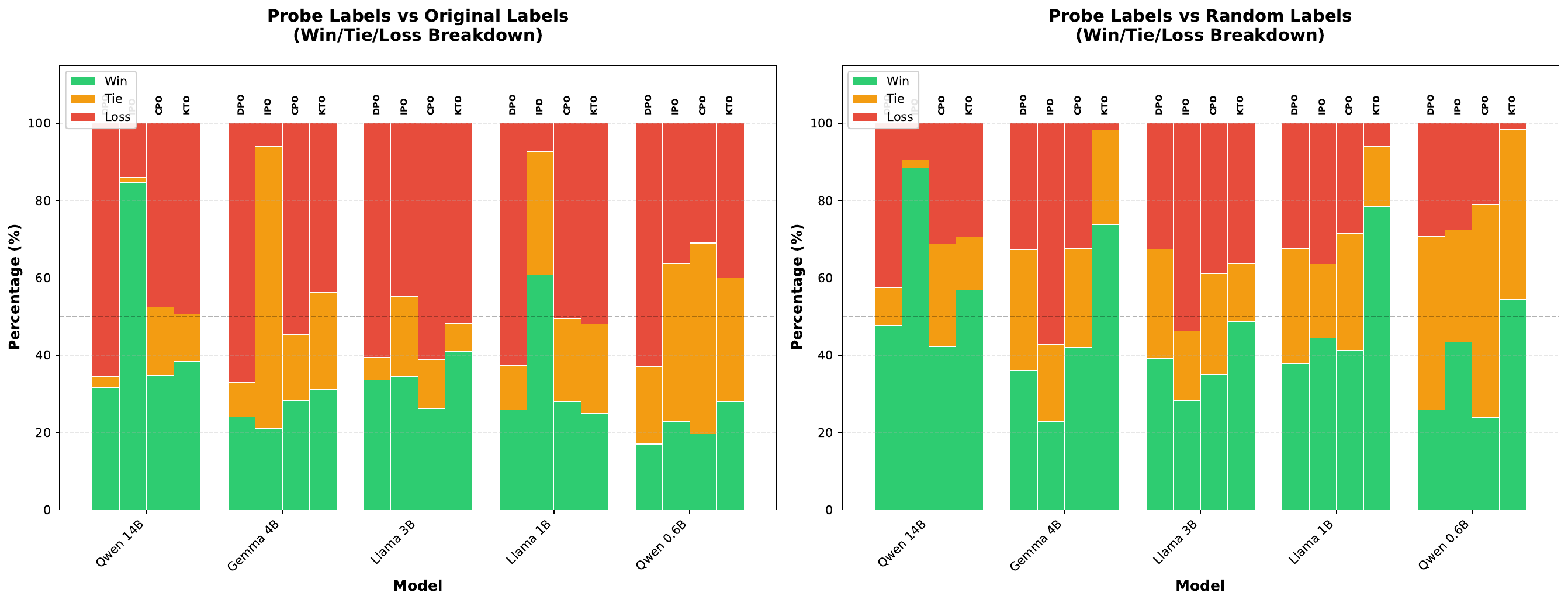}
    \caption{Visual summary of the results from Table~\ref{tab:main}: win$+$tie rates for probe-labelled models vs.\ original-labelled (left) and randomly-labelled (right) baselines on HH-RLHF, across all model families, scales, and preference optimisation methods.}
    \label{fig:main_results}
\end{figure*}

Figure~\ref{fig:main_results} provides a visual summary of the win$+$tie rates reported in Table~\ref{tab:main}. The figure makes the relative performance of each preference optimisation method (DPO, IPO, CPO, KTO) immediately apparent. Most strikingly, IPO with probe labels consistently reaches or exceeds the performance of models trained on 50K original labels across most model families and scales, while DPO shows the largest deficit. CPO and KTO occupy an intermediate position, where probe labels are clearly superior to random labelling but do not always close the gap to original labels.

\subsection{Probe Accuracy Across All Settings}
\label{app:probe_acc}

Table~\ref{tab:probe_acc_full} reports linear probe validation accuracy (with 95\% confidence intervals) across all combinations of model family and dataset, for both pretrained and PO checkpoints. Several observations are consistent with the analysis in Section~\ref{sec:geometry}:

\begin{itemize}
    \item The dataset is the dominant factor in probe accuracy. Nectar yields the highest accuracy across all model families ($\approx$75--80\%), while HH-RLHF is the most challenging ($\approx$58--62\%). UltraFeedback falls in between ($\approx$60--63\%). This ordering aligns with the visual cluster separation reported in Section~\ref{sec:geometry}.
    \item The difference between pretrained and PO checkpoints is small ($<$3\% in most cases), with some families (Gemma~3, Qwen~3) showing slightly higher PO accuracy and others (Llama~3.2) being essentially unchanged. This confirms that preference-relevant information is already present in the pretrained model's representations, making our pipeline applicable even before any alignment training.
    \item Confidence intervals are narrow ($\leq$4\%), indicating stable and reproducible probe accuracy across random seeds and training splits.
\end{itemize}

The HH-RLHF ceiling of approximately 60--62\% is consistent with the reported inter-annotator agreement of $\approx$63\% on that dataset, suggesting that our probe is approaching the fundamental noise floor of human preference annotation rather than being limited by model capacity or probe expressivity.

\begin{table}[h]
\centering
\caption{Linear probe validation accuracy (\%) with 95\% confidence intervals across model families and datasets. Results are reported for both the pretrained and PO model checkpoints. Each entry is averaged over three random seeds.}
\label{tab:probe_acc_full}
\adjustbox{width=\columnwidth}{
\begin{tabular}{llcc}
\toprule
Model & Dataset & Pretrained acc $\pm$ std (\%) & PO acc $\pm$ std (\%) \\
\midrule
Gemma 3 4B   & HH-RLHF       & 58.2 $\pm$ 1.8 & 57.7 $\pm$ 3.2 \\
Gemma 3 4B   & Nectar        & 77.7 $\pm$ 2.3 & 79.4 $\pm$ 1.5 \\
Gemma 3 4B   & UltraFeedback & 62.6 $\pm$ 3.3 & 62.8 $\pm$ 3.6 \\
Llama 3.2 3B & HH-RLHF       & 61.5 $\pm$ 1.9 & 61.0 $\pm$ 2.5 \\
Llama 3.2 3B & Nectar        & 79.0 $\pm$ 3.7 & 80.4 $\pm$ 3.3 \\
Llama 3.2 3B & UltraFeedback & 60.2 $\pm$ 2.4 & 62.3 $\pm$ 1.8 \\
Qwen 3 0.6B  & HH-RLHF       & 57.7 $\pm$ 1.6 & 59.4 $\pm$ 2.4 \\
Qwen 3 0.6B  & Nectar        & 75.3 $\pm$ 4.0 & 76.3 $\pm$ 2.5 \\
Qwen 3 0.6B  & UltraFeedback & 62.1 $\pm$ 3.1 & 60.0 $\pm$ 3.1 \\
\bottomrule
\end{tabular}}
\end{table}

\section{Optimised Probe Inference Algorithm}
\label{app:algorithm}

Section~\ref{sec:experiments} noted that label propagation, as a separate preprocessing step, adds approximately 70\% overhead to preference optimisation training time. This overhead can be entirely eliminated by fusing probe labelling with the preference optimisation forward pass, as described in Section~\ref{sec:experiments}.

Algorithm~\ref{alg:fused} presents a concrete implementation of this fused approach. The key insight is that the LLM forward passes required to compute the preference optimisation loss already produce the intermediate activations needed for probe scoring. By caching activations at layer $\ell$ during the standard forward pass and evaluating the linear probe (a single matrix--vector multiplication) before backpropagation, we add negligible computational cost while eliminating the separate labelling step entirely. The resulting algorithm is also online: labels are assigned dynamically each iteration, which could in principle enable the probe to be updated jointly with the policy, though we leave this direction to future work.

\begin{algorithm}[t]
\caption{Preference Optimisation with Optional Fused Probe Labelling}
\label{alg:fused}
\begin{algorithmic}[1]
\REQUIRE Policy $\pi_\theta$, dataset $\mathcal{D} = \{(x_i, y_i^a, y_i^b)\}_{i=1}^N$, PO method $\mathcal{L}_{\text{PO}}$ (e.g., DPO, IPO)
\REQUIRE \textit{Optional:} trained probe $f_\phi$, extraction layer $\ell$
\FOR{each training step}
    \STATE Sample batch $\mathcal{B} \subset \mathcal{D}$
    \FOR{$(x, y^a, y^b) \in \mathcal{B}$}
        \STATE $\mathbf{h}^a, \text{logits}^a \gets \textsc{Forward}(\pi_\theta, [x; y^a])$ \textcolor{gray}{\# Cache activations at layer $\ell$}
        \STATE $\mathbf{h}^b, \text{logits}^b \gets \textsc{Forward}(\pi_\theta, [x; y^b])$
        \STATE
        \IF{probe $f_\phi$ provided}
            \STATE \textcolor{gray}{\# Probe labelling --- negligible cost relative to LLM forward pass}
            \STATE $p^a \gets f_\phi(\mathbf{h}^a_\ell)$ \quad $p^b \gets f_\phi(\mathbf{h}^b_\ell)$ \quad \textcolor{gray}{\# Single linear layer evaluation}
            \STATE $(y^+, y^-) \gets (y^a, y^b)$ \textbf{if} $p^a > p^b$ \textbf{else} $(y^b, y^a)$
        \ELSE
            \STATE $(y^+, y^-) \gets$ original labels from $\mathcal{D}$
        \ENDIF
        \STATE
        \STATE Compute $\mathcal{L}_{\text{PO}}(\text{logits}^+, \text{logits}^-;\, \pi_\theta, \pi_{\text{ref}})$ \textcolor{gray}{\# Reuse cached forward pass results}
    \ENDFOR
    \STATE Backpropagate and update $\theta$
\ENDFOR
\end{algorithmic}
\end{algorithm}

\section{Additional Validation}
\label{app:additional_validation}

\subsection{Probe Calibration and Error Analysis}
\label{sec:error_analysis}

We analyse probe behaviour across the same nine (model, dataset) settings from Section~\ref{sec:geometry}, distinguishing \textit{pairwise} accuracy (used throughout the paper: which of two responses is preferred) from \textit{piecewise} accuracy (classifying a single response as chosen or rejected in isolation). The probe is a strong comparator but a poorly calibrated classifier: piecewise accuracy is 5--15 points lower than pairwise accuracy in the same setting. Confidence-based filtering yields only marginal precision gains, indicating that abstention trades away corpus size faster than it improves label quality — consistent with our decision to label every pair rather than filter by confidence (Section~\ref{sec:method}).

Manual error analysis on HH-RLHF and UltraFeedback shows failures concentrate on length-matched pairs, i.e., cases where the preference itself is most subjective; a manual check of the highest-confidence errors reveals no consistent pattern. On Nectar, errors instead show a length divergence between chosen and rejected responses, which we attribute to the dataset rather than to a probe-specific failure mode. In settings with substantial length imbalance, training the probe on a length-controlled subset may help avoid this bias; we leave this to future work.

\subsection{Human Evaluation}
\label{sec:human_eval}

Automatic LLM-as-a-judge evaluation, while convenient, carries its own preference biases~\citep{zheng2023judging}. To validate that our results are not an artefact of the judge model's own alignment, we conducted a small-scale human evaluation on HH-RLHF. Four non-author annotators were recruited on campus and evaluated a total of 250 response pairs across three model--method configurations: Gemma~3~4B / DPO, Llama~3.2~3B / DPO, and Llama~3.2~3B / IPO. Annotators operated during their paid work-time. For each pair, annotators were shown the prompt and two anonymised completions (one from the probe-labelled model and one from a baseline), and asked to select the better response or mark a tie. 

\begin{table}[h]
\centering
\caption{Human evaluation results (Win / Tie / Loss \%) for the probe-labelled model versus the Original (50K labels) and Random baselines on HH-RLHF. Each cell reports results from 250 comparisons across four annotators.}
\label{tab:human_eval}
\adjustbox{width=\columnwidth}{
\begin{tabular}{llcc}
\toprule
Model & Method & vs.\ Original (W / T / L) & vs.\ Random (W / T / L) \\
\midrule
Gemma 3 4B   & DPO & 21.7 / 17.4 / 60.9 & 42.0 / 30.0 / 28.0 \\
Llama 3.2 3B & DPO & 20.9 / 14.0 / 65.1 & 28.1 / 43.8 / 28.1 \\
Llama 3.2 3B & IPO & 76.2 /  4.8 / 19.0 & 52.6 / 21.1 / 26.3 \\
\bottomrule
\end{tabular}
}
\end{table}

The directional pattern in Table~\ref{tab:human_eval} is fully consistent with our automated evaluation reported in Table~\ref{tab:main}. Specifically:

\begin{itemize}
    \item \textbf{IPO with probe labels outperforms both baselines under human judgement.} Against the full 50K original-label baseline, the probe-labelled IPO model wins 76.2\% of comparisons---a striking result given the $100\times$ annotation advantage of the baseline. Against random labelling, it wins 52.6\% of comparisons.
    \item \textbf{DPO with probe labels trails the original-label baseline but clearly beats random labelling.} This mirrors the automated results: DPO is the most sensitive to label noise, but probe labels still provide a genuine signal above chance.
    \item \textbf{Human and automated evaluation agree directionally across all six comparisons.} This corroborates the use of LLM-as-a-judge as a reliable proxy at the scales required by Table~\ref{tab:main}.
\end{itemize}

We note that the sample sizes here reflect the practical constraints of human evaluation: a full-scale study covering the scope of Table~\ref{tab:main} (5 models $\times$ 4 methods $\times$ 2 baselines $\times$ 500 pairs) would require approximately 20,000 human judgements---precisely the annotation regime our method is designed to circumvent.

\subsection{Downstream Preference Optimisation on PRISM}
\label{sec:prism_po}

Section~\ref{sec:culturally_divers} demonstrated that the PRISM dataset~\citep{kirk2024prism} exhibits activation geometry analogous to standard preference datasets in the pretrained model, but that this geometry is substantially reduced or absent in default SFT/PO models. This finding strengthens the case for our pipeline in culturally specific settings: a pretrained (or SFT) model retains the representational structure needed to train a meaningful linear probe on population-specific labels, even if a publicly available PO model has obscured that structure through alignment to a different population's preferences.

\paragraph{Experimental protocol.}
Since LLM-as-a-judge evaluation would encode the preferences of a population different from the target group, we design an embedding-based evaluation protocol instead. We select the ``18--24 years old -- Africa'' subgroup as a representative low-resource minority group. We train the probe on 500 labelled samples from this subgroup and use it to annotate the full PRISM dataset ($\sim$27K pairs) for preference optimisation. As a fully-supervised baseline, we train directly on all available group-specific labelled samples.

For held-out samples from the target group, we generate completions from both models, embed them using \texttt{sentence-transformers/all-MiniLM-L6-v2} \citep{reimers2019sentencebert}, and compute cosine similarity to embeddings of the original labelled completions from that group. We report $\Delta = \mathrm{sim}_{\text{probe}} - \mathrm{sim}_{\text{original}}$; positive values indicate that the probe-trained model's generations are closer to the target group's preferred style.

\paragraph{Results.}

\begin{table}[h]
\centering
\caption{PRISM downstream experiment: difference in embedding similarity to target-group ground-truth completions between the probe-trained and fully-supervised models ($\Delta$ cosine similarity, with percentage change in parentheses). Positive values favour the probe-trained model.}
\label{tab:prism_po}
\adjustbox{width=\columnwidth}{
\begin{tabular}{lcccc}
\toprule
Model & DPO & IPO & CPO & KTO \\
\midrule
Gemma 3 4B   & $-0.006$ ($-3\%$)  & $+0.058$ ($+29\%$) & $+0.012$ ($+5\%$)  & $+0.027$ ($+12\%$) \\
Llama 3.2 1B & $-0.001$ ($-0.5\%$) & $+0.006$ ($+3\%$)  & $+0.023$ ($+10\%$) & $+0.013$ ($+6\%$)  \\
\bottomrule
\end{tabular}
}
\end{table}

In 7 out of 8 configurations, the probe-trained model produces completions that are more similar to the target group's ground-truth responses than the fully-supervised baseline, despite using only 500 labelled examples while the baseline uses all available group-specific data. IPO shows the most substantial improvement, with up to 29\% relative gain for Gemma~3~4B. This is consistent with IPO's noise robustness observed throughout the main experiments. DPO shows marginal degradation in both cases ($-3\%$ and $-0.5\%$), again mirroring its sensitivity to label noise in the main results.

We acknowledge that embedding-based cosine similarity is an imperfect proxy for human preference judgement, as it measures stylistic similarity to reference completions rather than subjective quality. Nevertheless, these results provide encouraging initial evidence that our pipeline transfers effectively to culturally specific, low-resource settings, and that the activation geometry documented in Section~\ref{sec:culturally_divers} translates into a usable preference signal for downstream adaptation.

\section{Hyperparameters}
\label{app:hyperparameters}

Table~\ref{tab:hparams} lists the hyperparameters used for all preference optimisation experiments. The learning rate was selected from a logarithmically spaced grid of $[5 \times 10^{-5},\, 5 \times 10^{-4}]$ using a held-out validation split of 500 examples. All other hyperparameters were fixed across all methods and model families to ensure a fair comparison. LoRA was applied to all attention projection matrices (query, key, value, output) with rank $r = 16$ and $\alpha = 32$. Label smoothing was set to 0.0 in all experiments except those in Table~\ref{tab:smoothing}, where it is explicitly varied.

\begin{table}[h]
\centering
\caption{Hyperparameters for all preference optimisation experiments.}
\label{tab:hparams}
\adjustbox{width=\columnwidth}{
\begin{tabular}{ll}
\toprule
Hyperparameter & Value \\
\midrule
Learning rate & $1 \times 10^{-4}$ (selected from $[5 \times 10^{-5},\, 5 \times 10^{-4}]$) \\
(Virtual) batch size & 64 \\
$\beta$ (DPO / IPO) & 0.1 \\
Label smoothing & 0.0 (unless otherwise stated) \\
Max sequence length & None for HH-RLHF; 4096 tokens otherwise \\
Training epochs & 1 \\
LoRA rank $r$ & 16 \\
LoRA $\alpha$ & 32 \\
LoRA target modules & All attention projections (Q, K, V, O) \\
Optimiser & AdamW ($\beta_1 = 0.9$, $\beta_2 = 0.999$, $\epsilon = 10^{-8}$) \\
Warmup steps & 100 \\
Infrastructure & A100 (80 GB) for training; A40 for evaluation \\
Training time & Up to 47h for 14B model. \\
\bottomrule
\end{tabular}}
\end{table}

\section{Potential Risks}

While our work is primarily foundational and designed to make preference alignment more accessible to underserved communities, we identify several potential risks associated with its use.

\paragraph{Amplification of Harmful Preferences.}
Our pipeline amplifies a small seed of labelled preferences into large-scale supervision. If the seed annotations reflect harmful, discriminatory, or otherwise undesirable preferences, the probe will propagate these at scale. This is particularly concerning in adversarial settings where a bad actor deliberately seeds the labelled set with, for example, preferences for toxic or misleading outputs. Practitioners should apply content filtering and careful auditing of seed annotations before deployment, and we recommend combining probe-based labelling with standard safety classifiers as a safeguard.

\paragraph{Misuse for Targeted Manipulation.}
The ability to cheaply adapt a model to population-specific preferences could be exploited to produce highly tailored disinformation or manipulative content. For instance, an actor could collect a small set of preference labels from a target demographic and fine-tune a model to generate text that resonates specifically with that group, facilitating microtargeted propaganda or influence operations. This dual-use risk is inherent to any label-efficient alignment method and is not unique to our approach, but the low annotation cost we achieve lowers the barrier to such misuse.

\paragraph{Exclusion and Bias Reinforcement.}
Although our method is motivated by the goal of serving non-mainstream populations, it could paradoxically reinforce exclusion. If the small labelled seed is unrepresentative of the full diversity within a target community---e.g., collected only from more accessible or vocal subgroups---the propagated labels will reflect the biases of that subgroup rather than the community at large. This risk is heightened for communities where internal diversity is high and where power imbalances may cause certain voices to dominate the annotation process. We encourage practitioners to invest in representative sampling strategies and participatory data collection even when seed sizes are small.

\paragraph{Stability and Misalignment at Small Scales.}
We document that our method becomes less reliable below 1B parameters. Practitioners deploying small models in resource-constrained settings---precisely the settings our method targets---should be aware that probe-based labelling can produce degenerate outputs at this scale, potentially causing unpredictable model behaviour in deployment. We recommend extensive evaluation before deploying probe-trained models below 1B parameters in user-facing applications.

\end{document}